%% file: main.tex
\documentclass[journal]{IEEEtran}
\usepackage{ifpdf}
\usepackage{booktabs}

\usepackage{cite}

\ifCLASSINFOpdf
  \usepackage[pdftex]{graphicx}
  \graphicspath{{../pdf/}{../jpeg/}}
  \DeclareGraphicsExtensions{.pdf,.jpeg,.png}
\else
  \usepackage[dvips]{graphicx}
  \graphicspath{{../eps/}}
  \DeclareGraphicsExtensions{.eps}
\fi
\usepackage{amsmath}
\usepackage{amsthm}
\usepackage{array}

\usepackage{epsfig}
\usepackage{graphicx}
\usepackage{tikz,graphics,color,fullpage,float,epsf,caption,subcaption}
\usepackage{fixltx2e}

\usepackage{stfloats}
\ifCLASSOPTIONcaptionsoff
  \usepackage[nomarkers]{endfloat}
 \let\MYoriglatexcaption\caption
 \renewcommand{\caption}[2][\relax]{\MYoriglatexcaption[#2]{#2}}
\fi
\usepackage{url}

\usepackage{algorithm}
\usepackage{algpseudocode}

\usepackage{dsfont}

\def\w{\mathbf{w}}

\def\1{\mathbf{1}}
\def\0{\mathbf{0}}

\def\optlimits{\nolimits}

\usepackage{tabu}

\usepackage{graphicx}
\usepackage{comment}
\usepackage{amsmath,amssymb} 
\usepackage{color}
\input{math_commands.tex}

\usepackage{bm} 

\usepackage[utf8]{inputenc} 
\usepackage{hyperref}       
\usepackage{url}            
\usepackage{booktabs}       
\usepackage{amsfonts}       
\usepackage{nicefrac}       
\usepackage{amsmath}
\usepackage{graphicx}
\def\optlimits{\nolimits}
\usepackage{dsfont}
\usepackage{verbatim}
\usepackage[sort, numbers]{natbib}

\def\w{\mathbf{w}}

\begin{document}
%
\title{Connectivity-constrained Interactive Panoptic Segmentation} 

%
%
%

\author{Ruobing~Shen,~\IEEEmembership{}
        Bo~Tang,~\IEEEmembership{}
        Andrea~Lodi,~\IEEEmembership{}
        Ismail~Ben~Ayed,~\IEEEmembership{}
        Thomas~Guthier~\IEEEmembership{}
\thanks{Email address: \texttt{rshen@g.clemson.edu}}}

%
%

\markboth{Journal of \LaTeX\ Class Files,~Vol.~14, No.~8, August~2015}%
{Shell \MakeLowercase{\textit{et al.}}: Bare Demo of IEEEtran.cls for IEEE Journals}
%



\maketitle

\begin{abstract}
We address interactive panoptic annotation, where one segment all object and stuff regions in
an image. We investigate two graph-based segmentation algorithms that both enforce connectivity of each region, with a notable  class-aware Integer Linear Programming (ILP) formulation that ensures global optimum.  Both algorithms can take RGB, or utilize the feature maps from any DCNN, whether trained on the target dataset or not, as input. We then propose an interactive, scribble-based annotation framework.
We present competitive semantic and panoptic segmentation results on the PASCAL VOC 2012 and Cityscapes  validation dataset given initial scribbles. We also demonstrate that our interactive approach can reach $90.6\%$ mIoU on VOC with just $3$ correction scribbles, and $62.8\%$ PQ on Cityscapes with $13$ minutes scribbles annotation plus one round ($2$ minutes)  of correction. 
\end{abstract}



%
\IEEEpeerreviewmaketitle

\section{Introduction}
%
%
%
%
\IEEEPARstart{D}{eep} Convolutional Neural Networks (DCNNs) excel at a wide range of image recognition tasks~\citep{7780459, 7485869, Shelhamer2017}, such as semantic segmentation~\citep{7913730, Shelhamer2017, zhao2017pspnet} and panoptic segmentation~\citep{2018arXiv180100868K,2019arXiv190205093Y, 2019arXiv190103784X, 2019arXiv190102446K}. Semantic segmentation studies the tasks of assigning a class label to each pixel of an image, while instance segmentation~\citep{Chen2018MaskLabIS} detects and segment each object instance. Panoptic segmentation unifies both tasks that investigate to segment both \emph{things} (such as person, cars) and  {stuff} (such as road, sky) classes.

While DCNNs show outstanding results for semantic and panoptic segmentation, they have two conceptional problems. First, they require huge amounts of annotated data. Annotating image segmentation masks is a very time consuming and labor extensive task. For example, annotating a semantic image mask took “more than 1.5h on average” on the Cityscapes dataset~\citep{Cordts2016Cityscapes}. 
In addition, DCNNs rely on their implicitly learned generalization probability and most of the state-of-the-art architectures do not make use of any domain specific knowledge, such as neighborhood relations and connectivity priors for segments. On the contrary, classical graph based segmentation models~\citep{969114, Yedidia:2003} do not require any learning data and can incorporate specific domain knowledge. Their major drawback is that they rely on human-designed similarity features and require complex optimization algorithms or solvers, which are mainly CPU based and  non-suitable for real time applications.

In this work, we explore the combination of DCNNs and graph-based algorithms for annotating ground truth segmentation. We investigate two interactive algorithms  that both enforce the \emph{connectivity prior} (to be more precise in Sec.~\ref{prerequisite}).
Specifically, we design a heuristic region growing method  based on the Potts model~\citep{potts_1952}, as well as an integer linear programming (ILP) formulation of the markov random field (MRF)~\citep{698673}. We utilize scribble based annotations from human annotators as initialized or iterative hard constraints for our  algorithms, which is typical in a human-in-the-loop (HITL) annotation process.
We explore two different scenarios. In the first scenario, we assume that for the segmentation task, there is already a pre-trained DCNN with the same class mapping. Thus, we could make full use of DCNN's probability map and scribbles as input to our algorithms.
In the second scenario, we assume that no pre-trained DCNN for the same objective is available. This is true for a lot of existing datasets, e.g., Cityscapes does not contain any class labels for lane marking, which is crucial information for an autonomous vehicle. In this case, we cannot use the class specific probability map, but the more generic low level features of a DCNN can be utilized as feature description for the algorithm. 

\begin{figure}[t]
\centering
\includegraphics[width=0.49\columnwidth]{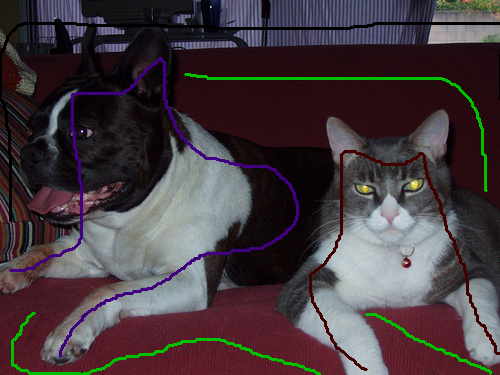}
\includegraphics[width=0.49\columnwidth]{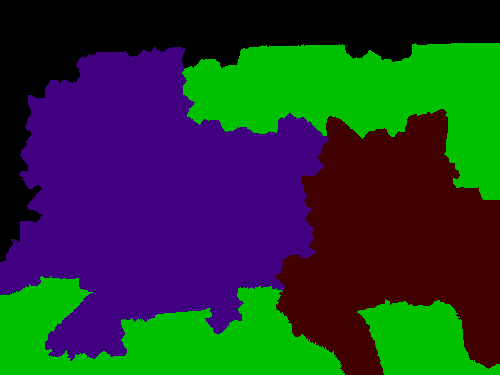}
\includegraphics[width=0.49\columnwidth]{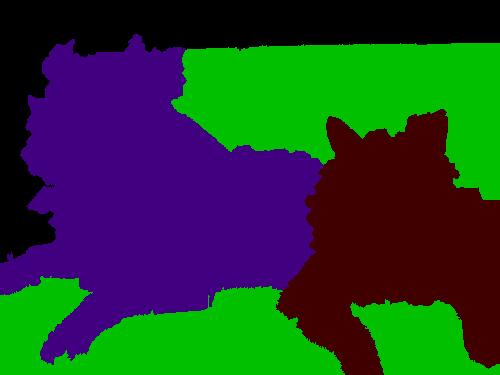}
\includegraphics[width=0.49\columnwidth]{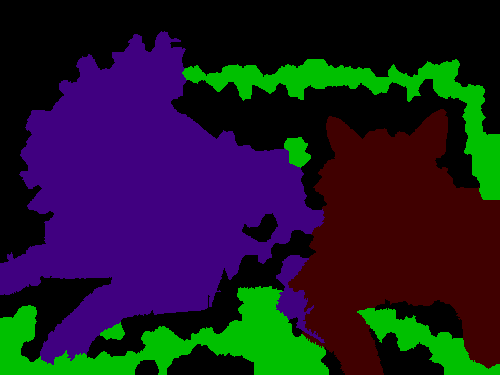}
\caption{ Top left: image with scribbles from Pascal VOC dataset. Top right: semantic segmentation result of our heuristic with layer 3 of ResNet $101$. Bottom left: result of our ILP with probability map of DeepLab V2. Bottom right: ILP without connectivity prior (with DeepLab V2 prob. map).}
\label{front_image}
\end{figure}

To investigate the general purpose of low level DCNN features, we compare DCNN trained on ImageNet~\citep{ILSVRC15} and COCO~\citep{10.100748} with and without fine tuning on the target dataset. Our experiments show that incorporating the connectivity prior as well as the DCNN features greatly improves the algorithm performance. We present competitive semantic (and panoptic) segmentation results on the PASCAL VOC 2012~\citep{Everingham15} and Cityscapes dataset.  To the best of our knowledge, our method is the first ``non-DCNN'' panoptic segmentation algorithm on Cityscapes with competing results, which shows the potential improvement gained by a combined approach. See Fig.~\ref{front_image} for a visualization that uses initialized scribbles as input.


We also prototype our algorithms for the task of interactive annotation. Based on initial scribbles and a baseline DCNN's probability map, our algorithms acheive $90.9\%$ mIoU on VOC validation set, with just $3$ rounds ($1$ scribbles per image each round) of correction scribbles.

Summarized, our key contributions are
\begin{itemize}
	\item a fast and good-performing heuristic algorithm and a ``global'' ILP formulation with connectivity prior, 
	\item an in depth analysis of a combination of DCNN and graph-based segmentation algorithms,
	\item extensive evaluation of the scribble based interactive algorithms for semantic and panoptic segmentations on two challenging datasets.
\end{itemize}

Our proposed algorithms have multiple use cases in annotating datasets for segmentation. First, 
they can be used inside any HITL annotation tool, as the annotator interacts with the image in forms of scribbles until satisfaction. Second, they can be used inside any weakly or semi-supervised learning framework for semantic (and panoptic) segmentation, which only requires initial scribbles~\citep{Lin2016ScribbleSupSC}. Third, semi-supervised learning in forms of scribbles can be combined with active learning framework~\citep{8215750}, to better utilize the huge amount of unlabeled data.

\paragraph{Related Work.}

The procedure of annotating per pixel segmentation masks is similar to interactive image segmentation, which is widely studied in the past decade. The method using bounding boxes~\citep{Rother:2004} is suitable for instance segmentation, which requires the user to draw the box as tight as possible. Recently, $4$ extreme points~\citep{Man+18, extremeclick} clicking is used as an alternative which shows superior result. 
On the other hand, polygon based methods~\citep{Russell2008} require users to carefully click the extreme points of things and stuff, and the accuracy heavily depends on the number of clicks.  Among others, scribbles are recognized as a more user-friendly way~\citep{937505, 4359322}. Moreover, it is also natural to annotate stuff classes using scribbles. 

Modern annotation tools often adopt interactive deep learning based methods, including Polygon-RNN++~\citep{acuna2018efficient} and  Curve-GCN~\citep{2019arXiv190306874L}, which allows the annotator to click on the boundaries.  Notably, a recent approach~\citep{agustsson2019interactive} enables both $4$ extreme points clicking and scribbles correction. Besides, they can also take advantage of ensemble learning~\citep{Zhou2009}, to combine several inference results to produce a better segmentation. However, these methods all require an existing ground truth dataset for DCNNs to learn on the first hand, which may not be available when unknown domains or new classes are introduced.

As a cheaper alternative, weakly supervised learning has drawn a lot of attention recently. ~\citep{Dai2015BoxSupEB, khoreva_CVPR17, Li_2018_ECCV} claim that weakly iteratively trained by just bounding boxes and image tags, the DCNN can achieve $95\%$ segmentation score compared to fully supervised on VOC. Since this method emphasizes on thing classes, it has worse score (or even none) on stuff classes.  Instead, ~\citep{Lin2016ScribbleSupSC, 20.500} claim that iteratively training a DCNN by scribble annotations suffers only a small degradation in performance on both thing and stuff classes.

For graph-based methods, the (discrete) Potts model~\citep{potts_1952} is widely used for denoising and segmentation. The authors of~\citep{DBLP-1803-07351} formulate the problem as an ILP and try to solve the global optimum, but only to a reduced image size. The authors of~\citep{7410389} proposed an efficient region-fusion-based heuristic algorithm, while~\citep{rshen_ilp2017} extends the work to allow scribbles as input.
The MAP-MRF (maximizing a posterior in an MRF) has been well studied for image segmentation. Previous methods focus on local priors \citep{Boykov2006, NIPS2011_4296}, and efficient approximate algorithms exist, e.g., graph cut~\citep{969114} and belief propagation~\citep{Yedidia:2003}. MRF with connectivity prior has been studied in~\citep{Rempfler2016TheMC, 5828, 4587440} and formulated as an ILP, which can be solved by any ILP solver~\citep{doi:10.1080}. ILP with root or scribble nodes connectivity was studied in~\citep{rshen_ilp2017, Rempfler2016TheMC, rshen2018}, where for each class at least one node is fixed.
Since solving an ILP exactly is in general $\mathcal{NP}$-hard~\citep{Land60anautomatic}, they either solve a relaxation of the ILP, or focus on the binary MRF case.

Our interactive (in forms of scribble) multi-label annotation algorithms are graph based, with global connectivity prior, i.e., they enforce pixels of the same label to be connected, which allows the annotator to better control over the global segmentation.

\section{Proposed approach}

\subsection{Prerequisite}
\label{prerequisite}
Given an image, we build an undirected graph $G=(V,E)$ where the node set~$V$ represents a set of pixels (or superpixels)  and~$E$ a set of edges consisting of un-ordered pairs of nodes.
Image segmentation can be transformed into a graph labeling problem, where the label set $C$ is pre-defined.

When talking about segmentation, we need to first  distinguish between class, instance, and region ID of a node. In semantic segmentation,  the task is to assign a class label to each node in a graph. In panoptic segmentation, one has to further assign an instance ID to the node that belong to the ``thing'' class. In this paper, our algorithms require an additional region ID, which is linked to a scribble and we assume nodes with the same region ID must be connected (to be explained in Sec.~\ref{extract_scribble}). This is to deal with the case where an object of the `thing'' or ``stuff'' class is separated into several connected regions, e.g., the car in Fig.~\ref{scribble} is separated into two regions by a tree. Afterwards, the class and region labels can be used to generate a panoptic segmentation.

\subsection{Connectivity-constrained optimization algorithms}
\label{proposed}
We first discuss the formal definition of connectivity, and the two proposed algorithms in details, i.e., the class-agnostic heuristic algorithm of the discrete Potts model and the class-aware ILP of MRF with connectivity constraints.
\subsubsection{The connectivity prior}
\label{connectivity}
Two nodes $u,v$ in a graph $G$ are \emph{connected} if there is a $(u,v)$-path in $G$.
$G$ is called \emph{connected} if every pair of nodes are connected in $G$, otherwise it is \emph{disconnected}. Let $\bar{G_\ell} \subseteq G$ be a connected subgraph where every node is labeled $\ell\in C$. Then, the image segmentation with connectivity constrains corresponds to find a partition of $G$ into $k$ ($k = |C|$) connected (and disjoint) subgraphs.
Enforcing connectivity constraints itself is proven to be $\mathcal{NP}$-hard in~\citep{5828}. \citep{rshen_ilp2017} fixed at least one node of each subgraph by scribbles. However, the constraints are still $\mathcal{NP}$-hard.

\subsubsection{The \texorpdfstring{$\ell_0$} region fusion based heuristic}
\label{l0heur}
Given a graph $G(V,E)$, let $y_i$ be the information (either RGB or features from any DCNN) of node~$i$, and $w_i$ be its estimated value, the discrete Potts model~\citep{potts_1952} has the following form:
\begin{equation}
\label{discrete_potts}
\min_{\w} \sum\optlimits_{i\in V}\left\Vert w_i - y_i \right\Vert_2 + \sum\optlimits_{(i,j)\in E}\lambda \left\Vert w_i - w_j \right\Vert_0 ,
\end{equation}
where $\lambda$ is the regularization parameter. Here, the first term is the data fitting and the second is the regularization term. We recall that the $\ell_0$ norm of a vector gives its number of nonzero entries.

In this paper, we introduce an iterative scribble based region fusion heuristic algorithm (which we call $\ell_0H$) with the ``class'' and ``region'' ID for each node.
In the beginning, the nodes covered by the same scribble are grouped together and labeled with the same IDs, while  all other nodes are unlabeled and in their individual group. Note that different regions can share the same class ID. Then, the algorithm iterates over each group (outer loop) and its neighbors (inner loop) and decide whether to merge the neighbors or not, by checking their region IDs. If both groups have region IDs and are different, they cannot be merged. In all other cases, i.e., if both groups have no region ID or only one has it, the following merging criteria~\citep{7410389} are checked:
\begin{equation}
\label{lzero}
 \sigma_i\cdot \sigma_j\cdot\left\Vert Y_i - Y_j \right\Vert_2 \leq \beta\cdot \gamma_{ij}\cdot(\sigma_i+\sigma_j).
\end{equation}
where $\sigma_i$ denotes the number of pixels in group $i$, $Y_i$ the mean of image information (e.g., RGB color) of group $i$, and $\gamma_{ij}$ denotes the number of neighboring pixels between groups $i$ and $j$. Here,~$\beta$ is the regularization parameter, and it increases over the outer loop number.

If~(\ref{lzero}) is satisfied, two groups are merged, and their labels are updated according to the following rule. If both groups have no ``region'' ID, the merged group still have none, hence unlabeled. If only one group has region ID, the merged group inherits the label, hence labeled.

After one outer loop over all groups, $\beta$ will increase and follows the exponential  growing strategy of~\citep{7410389}, i.e., $ \beta = (\frac{\text{iter}}{50})^{2.2}*\eta$, where ``iter'' is the current number of outer loop and~$\eta$ is the growing parameter. The procedure goes on until all groups are labeled, and the complexity of $\ell_0H$ is $\mathcal{O}(n)$, where $n$ is the number of nodes.

Note that the above algorithm is approximate to problem~(\ref{discrete_potts}), and connectivity of each region is enforced at every step. Given desired scribbles, $\ell_0H$ is able to generate panoptic segmentations (and also semantic segmentations). Also note that the class ID does not play any role in the algorithm, it inherits from the scribble and propagates with the region ID. Hence, this algorithm is class-agnostic.





\subsubsection{The ILP formulation with connectivity constraints}
The MRF with pairwise data term can be formulated as  the following ILP:
\begin{align}
\begin{split}
\text{min}_{x} & \sum\optlimits_{\ell \in C}\sum\optlimits_{i\in V} c_i^\ell x_{i}^\ell \\
+ & \lambda \sum\optlimits_{\ell \in C}\sum\optlimits_{(i,j)\in E}d_{ij}|x_i^\ell-x_j^\ell| 
\end{split} \label{ILP} \\
\sum\optlimits_{\ell \in C}x_{i}^\ell&=1, \;\;\forall i\in V, \label{ILP1}\tag{\ref{ILP}a} \\
x_{i}^\ell&\in\{0,1\}, \;\;\forall i\in V, \;\; \ell \in C, \tag{\ref{ILP}b}
\end{align}
where $c_i^\ell$ denotes the unary data term for assigning class label~$\ell$ to node~$i$ (hence class-aware),  $d_{ij}$ the simplified pairwise term for assigning $i$, $j$ different labels, and~$\lambda\ $ is the regularization parameter. 
Constraint~(\ref{ILP1}) enforces that to each node is assigned exactly one label, i.e., $x_i^\ell = 1$ if and only if node $i$ is labeled~$\ell$. Note that the absolute term can be easily transformed into linear terms by introducing additional continuous variables. 
The model above does, however, not guarantee connectivity, which is instead enforced as follows. 

\paragraph{Connectivity constraints with root node}
Let $r$ (the first node of scribble $\ell$) denote the root node of subgraph $\bar{G_\ell}$. Then, the following constraints of~\citep{Rempfler2016TheMC} suffice to characterize the set of all connected subgraphs of label $\ell$  that contain $r$
\begin{equation}
x^\ell_i \leq \sum\optlimits_{s\in {S}} x^\ell_s, \;\;\forall  i\in V\backslash r: (i,r)\notin E, \; \forall S\in \mathcal{S}(i,r), \label{root}
\end{equation}
where $S$ is the \emph{vertex-separator set} of $\{i,r\}$, i.e., if the removal of $S$ from $G$ disconnects $i$ and $r$. And $\mathcal{S}(i,r)$ is the collection of all vertex-separator sets of $\{i,r\}$.

The number of constraints~(\ref{root}) is exponential with respect to the number of  nodes in $G$, hence it is not possible to include into problem~(\ref{ILP}) all of them at in once. In practice, they are added iteratively when needed (called the cutting planes method~\citep{doi:10.1137}). In this paper, we adopt the \emph{K-nearest cut generation} strategy~\citep{Rempfler2016TheMC} to add a few of them at each iteration. See appendix for more details.

\begin{figure}[t]
\centering
\includegraphics[width=0.9\columnwidth]{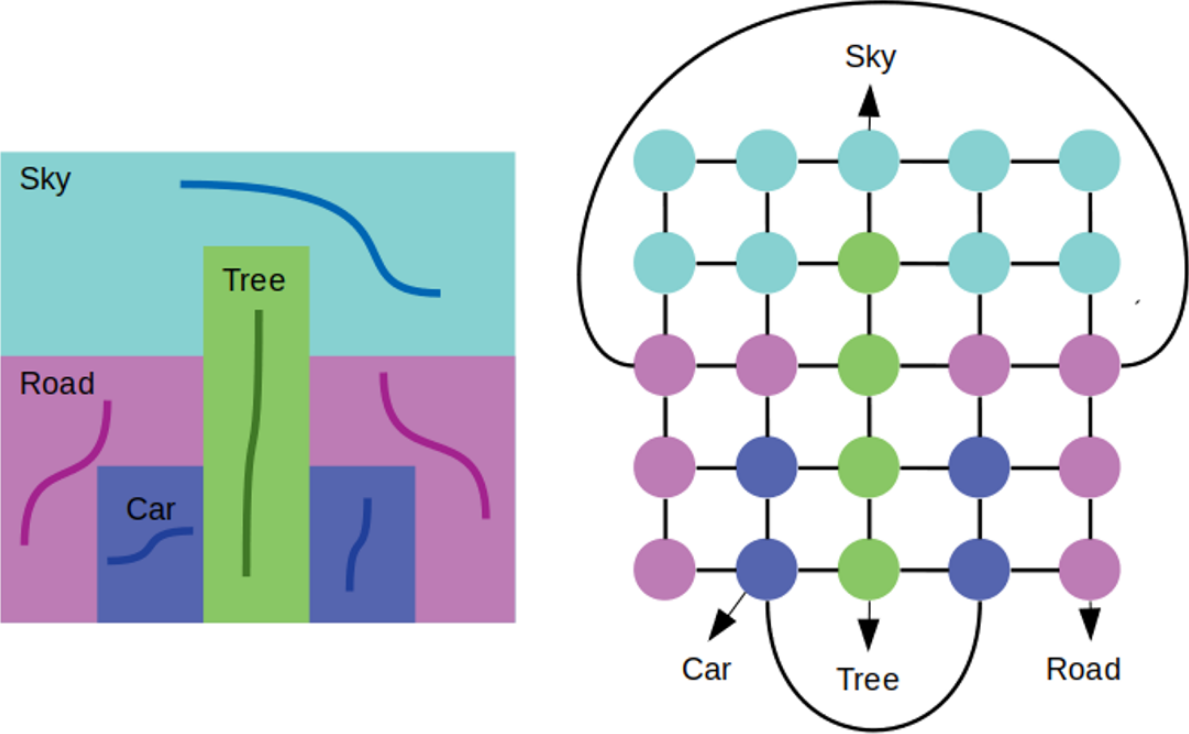}
\caption{ Left: our scribble policy to draw as many scribbles as there are separated regions. 
Right: ILP-P that introduces pseudo edges which connect multiple regions of the same class.}
\label{scribble}
\end{figure}

For the simple case where the region ID coincides with the class ID, i.e., the number of regions equals that of classes, problem~(\ref{ILP}) with connectivity prior is solved as follows. We first solve~(\ref{ILP}) and check if all subgraphs $G_i$ are connected. If not, we identify disconnected components, adds constraints~(\ref{root}), and solve the resulting ILP again. This procedure continues until all subgraphs are connected. This method ensures global optimality if no time limit is restricted.



We introduce an ILP formulation (we call it ILP-P, P for Panoptic) where we add $k-1$ pseudo edges that ``connect'' all separated regions of the same class (illustrated in Fig.~\ref{scribble}). In particular, this does not increase the number of variables and is class-aware. Then, the solving process follows the aforementioned framework. But ILP-P is only designed for semantic segmentation, i.e., it is not region-aware. Post-processing methods are needed to separate instances within the same class for panoptic segmentation.


\section{Overall workflow}
In this section, we describe the pre-processing steps as well as the scribble policy for our panoptic and interactive segmentation.

\subsection{Superpixels as dimension reduction}
\label{superpixel}
Superpixels have long been used for image segmentation~\citep{STUTZ2017, SLIC, DBLP-1803-07351}, as they can greatly reduce the problem size while not sacrificing much of the accuracy. In this paper, we adopt SEEDS~\citep{seeds} to generate superpixels on the PASCAL VOC 2012 dataset, while using a deep learning based method~\citep{Tu-CVPR-2018} on the more challenging Cityscapes dataset. We then build a region adjacency graph (RAG) $G(V,E)$  of the superpixels, where each superpixel forms a node (vertex) and edges connect two adjacent superpixels. 

\subsection{Extracting image features using scribbles}
\label{extract_scribble}
Our segmentation algorithms are scribble supervised, which are two folded. On the one hand, the node labels, such as class, instance and region IDs are fixed if the nodes are covered by the corresponding scribbles. On the other hand, for the ILP algorithm, if no high level image information (i.e., probability map) exists, the scribbled superpixels will be used to extract information for the class, i.e., one can use the average of the scribbled superpixels information  to represent each class. 

Although the input to the algorithms can be as simple as image RGB information, one can also take advantage of the modern DCNN to extract deeper features. We distinguish two scenarios:
\begin{itemize}
    \item No previous training data is available -- one starts annotating images in a new dataset.
    \item Training data available -- one continues annotating more images of an existing dataset.
\end{itemize}
In the former case, other than RGB, one can also adopt any base network (i.e., ResNet $101$~\citep{7780459}) pre-trained on other datasets (i.e., ImageNet) and use the output of the low level features that extracts image edges, textures, etc.
In the later case, one can fully utilize any modern DCNN trained on the existing dataset, and use the output of the final layer (i.e., probability map).

\paragraph{Scribble generation policy}
For panoptic segmentation, first of all, the scribble itself must be connected. Second, one has to draw as many scribbles as there are connected regions (both ``thing'' and ``stuff'' class) presented in the image. For example, if an object is cut into separated regions, one has to draw a scribble on each region. One sample image with scribbles is shown in Fig.~\ref{scribble}.

\subsection{Interactive segmentation using scribbles}
In a HITL annotation framework, it is typical that the annotator makes corrections to the current segmentation until satisfaction.
In this paper, we simulate the human correction scribbles (described in Appendix) and manually correcting annotations~\footnote{Both the simulating code and manual labels will be open source upon acceptance.}. We then report analysis of the interactive segmentation results  in Sec.~\ref{interactive_exp}.

\section{Semantic and Panoptic Segmentation Experiments}
\label{experiments}

\subsection{Experimental setup}
In this session, we conduct extensive experiments on the Pascal VOC 2012 and Cityscapes validation set.
In all our experiments, when we mention base network, we refer to the publicly released ResNet $101$ that is pre-trained on ImageNet and COCO dataset. We adopt DeepLab V2~\citep{7913730} (without CRF as post-processing) and DRN~\citep{Yu2017} as our baseline DCNN to get the probability maps, trained on their corresponding training sets.
We adopt IBM Cplex~\citep{Bliek2014SolvingMQ} version $12.8$ to solve the ILP.
All computational experiments are performed on a Intel(R) Xeon(R) CPU E$5$-$2620$ v$4$ machine, with $64$ GB memory.

We report the semantic and panoptic segmentation scores, where the pixel mean intersection over union (mIoU) is commonly used for semantic segmentation, and the panoptic quality (PQ) metric is newly introduce in~\citep{2018arXiv180100868K} and is a combination of segmentation quality (SQ) and recognition quality (RQ).

\paragraph{Parameter settings for the experiments}
\label{parameter}
Parameter $y_i$ is the information contained in pixel $i$, which could come from a vector of RGB channels, layer $1$ or layer $3$ feature of ResNet $101$, or from the probability map of a trained DCNN. For VOC 2012, the probability map comes from a trained DeepLab V$2$. For Cityscapes, we use DRN to get the probability map inference. 

For $\ell_0H$, after several trials, the growing parameter $\eta$ is set to $0.1$, $20$, $100$ and $0.3$ for RGB, layer $1$, layer $3$ and probability map of both VOC 2012 and cityscapes. 
For ILP, when training data is available, we can use the probability map $p_i$ and $c_i^\ell = \left\Vert \mathds{1}^\ell - p_i \right\Vert_2$, where $\mathds{1}^\ell$ is an $k$ ($k$ being the number of classes) dimensional vector with  $\left\Vert \mathds{1}^\ell \right\Vert_1=1$ and the $\ell$'s position equals $1$. When there is no training data, we compute the average of the nodes information ($y_i$) covered by scribbles of the same class (i.e., class $\ell$), and use this  to represent class $\ell$ (denote as $Y_\ell$). Then $c_i^\ell =\left\Vert y_i - Y_\ell \right\Vert_2$. The regularization parameter $\lambda$ for all the ILP is set to $100$, and the pairwise term $d_{ij} = e^{-\left\Vert y_i - y_j \right\Vert_2}$.

\subsection{Results on Pascal VOC 2012}
Pascal VOC 2012 has 20 “thing” classes and  a single “background” class for all other classes. We evaluate our algorithms on the $1449$ validation images.
We first apply~\citep{seeds} to produce around $700$ superpixels, and use the public scribbles set provided by~\citep{Lin2016ScribbleSupSC} as initial scribbles.
Since some of these scribbles do not meet our policy (described in Sec.~\ref{extract_scribble}), hence only the semantic segmentation is reported, in terms of mIoU averaged
across the $21$ classes. 

\paragraph{No training data } 
In this case, one can either use RGB or the output of lower level features of a base network as input ($y_i$) to our algorithms.
We compare our class-agnostic heuristic ($\ell_0H$) using RGB or different low level features from ResNet $101$ as input, against ILP-P.
We do not set any time or step limit for the heuristic, but a time limit of 10 seconds for ILP-P (denoted ILP-P-10). We report in Table~\ref{heuristic-table} the detailed comparison, where we use the RGB, first and third layer of ResNet $101$ as input to $\ell_0H$, and ``Dim'' is the dimension
of the input feature map.

\begin{table}[t]
    \caption{Comparison on VOC 2012 \emph{val} set when no training data is available.}
    \label{heuristic-table}
    \centering
        \begin{tabular}{llll}
        \toprule
        Model &   Dim   & Time     & mIoU \\
        \midrule
        $\ell_0H$-RGB   & 3 & 2.2 &  69.8   \\
        $\ell_0H$-layer 1   & 64  & 2.9 & 70.8      \\
        $\ell_0H$-layer 3    & 256 &    3.9    & 71.6  \\
        ILP-P-10 & -- &     9.7    & \textbf{71.9}  \\
        \bottomrule
    \end{tabular}
    \caption{Comparison when training data available, baseline DeepLab V$2$~\citep{7913730}.
    }
    \label{ILP-table}
    \centering
    \begin{tabular}{lll}
        \toprule
        Model&  Time  (s)     & mIoU ($\%$) \\
        \midrule
        DeepLab V$2$  & -- & 70.5     \\
        ILP-U &  0.2    &    80.8      \\
        $\ell_0H$-prob & 0.8 &    81.7     \\
        ILP-P-10  &  7.9    &    \textbf{84.6}      \\
    \bottomrule
  \end{tabular}
\end{table}

We can see in Table~\ref{heuristic-table} the advantage by incorporating lower level features maps of ResNet $101$, that improves $\ell_0H$ of RGB by  $1.8\%$, even though it is pre-trained on completely different dataset. ILP-P adopts $\ell_0H$-layer $3$ as initial solution, and further increase the mIoU by $0.3\%$.


\paragraph{With training data}
If training data is available, we use the probability map of DeepLab V$2$ (with baseline mIoU $70.5\%$) trained on VOC training set as input to our algorithms, $\ell_0H$ get another huge boost of $10.1\%$ to $81.7\%$ compared to using layer $3$.
We compare ILP~(\ref{ILP}) without connectivity (denoted ILP-U, U for unconnected) that is solved to optimum  with $\ell_0H$ and found out it is $0.9\%$ worse, which shows the importance of the connectivity prior. 
We then compare ILP-P with $\ell_0H$ as initial solutions ($81.7\%$ mIoU) and a time limit of $10$ seconds, Table~\ref{ILP-table} suggests that 
ILP-P-$10$ improves the baseline of DeepLab V$2$ by $14.1\%$ and $\ell_0H$-prob by $2.9\%$. 
An example of visual comparison is illustrated in Fig.~\ref{front_image}.

\subsection{Results on Cityscapes}
Cityscapes has 8 “thing” classes, and 11 “stuff” classes. 
We evaluate our algorithms on the $500$ validation images with the scribbles, where we first apply~\citep{Tu-CVPR-2018} to produce exactly $2000$ and $4000$ superpixels, which we call sup-2000 and sup-4000. If not otherwise specified, sup-2000 is by default.

\subsubsection{Simulated and manual scribble annotation}
\label{scribble_gen}

Since there is no public scribble set for Cityscapes, we first ``hack'' the ground truth instance segment and apply erosion and skeleton algorithms to simulate scribbles of $500$ validation images. We then adopt our own annotation tool to manually annotate $100$ of them, which takes roughly $13$ minutes per image.  Due to resource and time constraints, we only annotate 19 classes, and draw an ``ignore'' scribble on all other classes. 
And both scribbles follow that of Cityscape coarse annotations.

Note that we ensure both types of scribbles meet our policy in Sec.~\ref{extract_scribble}. Besides, their formats comply with Cityscapes that have both class and instance labels.
One can see Fig.~\ref{city_image} as an example of scribbles. 

\subsubsection{Results on simulated scribble annotation}
\paragraph{No training data } 
We use RGB and output of layer $3$ of Resnet $101$ as input to $\ell_0H$, and report in Table~\ref{heuristic-city} both mIoU and PQ. 
We also compare $\ell_0H$  to one recent weakly supervised learning method~\citep{Li_2018_ECCV}. It uses a more powerful PSPNet~\citep{zhao2017pspnet} supervised by bounding boxes and image tags, but requires end to end training. $\ell_0H$ shows superior results on both semantic and panoptic segmentation, both around $10\%$ boost compared to~\citep{Li_2018_ECCV}. Note that it is not a fair comparison since ours requires scribbles while~\citep{Li_2018_ECCV} does not at inference time.


\paragraph{With training data}
We use the public full-supervised (trained on Cityscapes) DRN~\citep{Yu2017} as our baseline ($71.4\%$ mIoU), and run 
$\ell_0H$ and ILP-P using its probability map. Table~\ref{heuristic-city} shows $\ell_0H$-prob improves the baseline by $9.4\%$, and by $6.5\%$ compared to $\ell_0H$-layer $3$ . Furthermore, PQ also  increases from $49.6\%$ to $61.4\%$. We further test $\ell_0H$ based on  $4000$ superpixels, and mIoU and PQ both increase another $1.4\%$ and $2.9\%$, with PQ scoring $64.3\%$.

We then look into the ILP formulation with and without connectivity.
Note that during computation, we ignore superpixels that are surrounded by scribbles, and we also cut the superpixel into smaller ones if touched by multiple  scribbles.
The average number of resulting superpixels (nodes) is $2170$ for sup-2000 and $3822$ for sup-4000.

\begin{table}[t]
    \caption{Comparison on Cityscapes \emph{val} set based on simulated scribbles, with and without its training data set. The later is based on DRN~\citep{Yu2017}'s probability map.  }
    \label{heuristic-city}
    \centering
        \begin{tabular}{lllllll}
        \toprule
        Model&  Time &  mIoU   & PQ & SQ  & RQ\\
        \midrule
        $\ell_0H$-RGB   &  6.5  &74.2   & \textbf{49.6} &74.3 &63.8 \\
        $\ell_0H$-layer 3  &7.2     & \textbf{74.3} &  \textbf{49.6}& 74.5 & 63.7  \\
        Weakly~\citep{Li_2018_ECCV} & --       & 63.6  &  40.5& -- & --\\
        \midrule
        DRN (baseline)   & -- & 71.4   & -- & --& -- \\
        $\ell_0H$-prob  & 5.1 &   80.8  & 61.4 & 76.3 & 78.7  \\
        $\ell_0H$-prob-4000  & 7.0 &   82.2  & \textbf{64.3} & 76.6 & 82.6  \\
        ILP-U-4000  & 1.5 &   \textbf{82.7}  & -- & -- & --\\
        ILP-P-20  & $20^*$ &   81.0  & 62.3 & 76.4 & 80.0\\
        \bottomrule
        \end{tabular}
\end{table}

Due to the huge number of integer variables (number of nodes multiply number of classes), ILP-P could not find any better solutions within $20$ seconds time limit in most of the cases, given initial solution of $\ell_0H$-prob. (Note that Cplex applies pre-processing that we can not set time limit on, hence we report $20$ seconds as average computation time.) However, it is still able to discover $75$ optimal solutions out of $500$ images. It is also surprising that in most of the $75$ cases,the optimal solution was found within $1$ second. The average mIoU is $81.0\%$, which is $0.2\%$ improvement compared to $\ell_0H$-prob.
On the other hand, ILP-U could find optimal solutions of all $500$ images in short time, averaging only $1.5$ seconds on sup-4000. It achieve the best result of $82.7\%$, which is $0.5\%$ better than $\ell_0H$-prob-4000, and $0.5\%$ better than ILP-P.

Although ILP-U performs best on mIoU score, its result does not ensures connectivity. This causes two drawbacks. Firstly, pixels of each class could appear anywhere in the image, which requires the annotator erasing mislabeled pixels all over the image. Secondly, one can not generate panoptic segmentation based on ILP-U, because of the same reason. 
On the contrast, based on the result of ILP-P, one can simply build a graph on those nodes with the same class and apply any post-processing algorithm to separate instances based on the scribbles. 

In this paper, based on the semantic segmentation of ILP-P, we adopt $\ell_0H$ to produce panoptic segmentation. The PQ results are $0.9\%$ better than that of $\ell_0H$-prob. Yet, there are many ways to boost the performance by applying more sophisticate post-processing. For example, given any edge detector network, one could apply another ILP-P to get instance segmentation. This will be leaved for further research. An illustration can be seen in sixth row of Fig.~\ref{city_image}.

\begin{table}[t]
    \caption{Experiments on Cityscapes \emph{val} set with manual scribble annotation.  }
    \label{manual-city}
    \centering
        \begin{tabular}{lllllll}
        \toprule
        Model&  Time &  mIoU   & PQ & SQ  & RQ\\
        \midrule
        DRN (baseline)   & -- & $71.1^*$   & -- & --& -- \\
        $\ell_0H$-prob  & 5.0 &   78.4  & 58.5 & 77.2 & 73.9  \\
        $\ell_0H$-prob-4000  & 6.7 &   79.9  & \textbf{61.5} & 77.6 & 77.7  \\
        ILP-U  & 0.6 &   \textbf{80.0}  & -- & -- & --\\
        ILP-P-20  & $20^*$ &   78.7  &  58.3& 77.2 & 73.7\\
        \bottomrule
    \end{tabular}

    \caption{Experiments on Cityscapes with 2 minutes modification scribbles.  }
    \label{modi-city}
    \centering
        \begin{tabular}{llllll}
        \toprule
        Time &  mIoU   & PQ & SQ  & RQ\\
        \midrule
         -- & $71.1^*$   & -- & --& -- \\
         4.9 &   79.1  & 59.3 & 77.4 & 74.9  \\
         6.6 &   \textbf{80.5}  & \textbf{62.8} & 77.7 & 79.5  \\
         0.6 &   80.1  & -- & -- & --\\
         $20^*$ &   79.4  & 59.2  & 77.3 & 74.8 \\
        \bottomrule
    \end{tabular}

\end{table}

\subsubsection{Results on manual scribble annotation}
In order to verify the usefulness of our algorithm in practice, we conduct similar experiments on the $100$ human annotated images. Although not as good as those based on ground-truth simulated scribbles, results are on par with the previous experiment. Keep in mind that we did not zoom in the image to fine annotate the far away instances, where might be some annotation errors. Besides, we compute the DRN~\citep{Yu2017}'s semantic segmentation on these $100$ images, and the mIoU is $71.1\%$, which means these annotated $100$ images are harder  than average.

 Both ILP~(\ref{ILP}) and $\ell_0H$ is still able to improve largely based on the DCNN, i.e., $8.8\%$, $8.9\%$ and $7.6\%$ for 
$\ell_0H$-4000, ILP-U and ILP-P. And ILP-U still performs best ($80.0\%$) in terms of mIoU, without the guarantee on connectivity. Please refer all the results in Table~\ref{manual-city}. One example could be seen in Fig.~\ref{city_image}, that a "Person" come out nowhere in ILP-U (in red color).

Finally, since ILP~(\ref{ILP}) is class-aware and encodes pairwise term $d_{ij}$, further boost of performance on semantic and panoptic segmentation can be expected, given better baseline DCNN and an edge detector.

\subsection{Results on interactive segmentation}
\label{interactive_exp}

\subsubsection{VOC2012}
\begin{figure}[t]
\centering
\includegraphics[width=0.99\columnwidth]{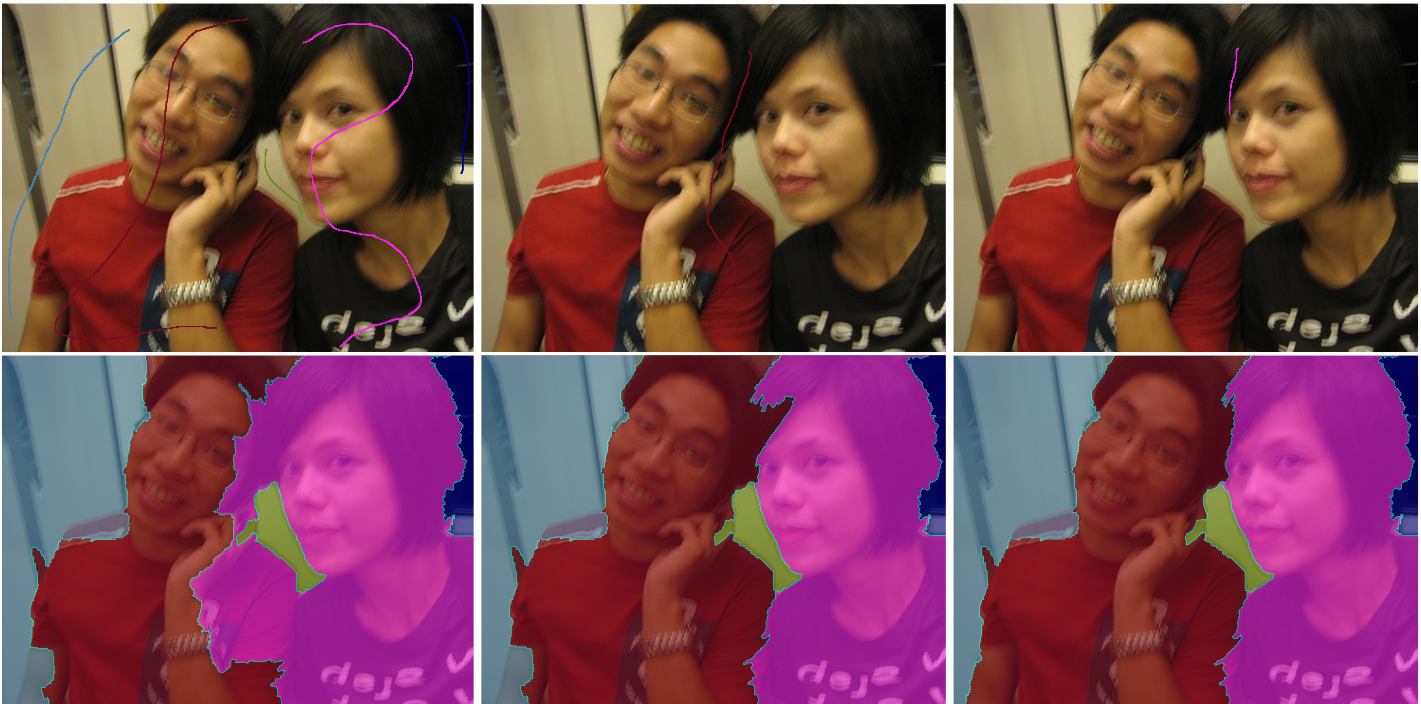}
\caption{Top: Images from VOC with initial scribbles and $2$ rounds of  scribbles correction. Bottom: Their corresponding panoptic segmentation result.}
\label{scribbles}
\end{figure}

\begin{figure}[ht]
\centering
\includegraphics[width=0.9\columnwidth]{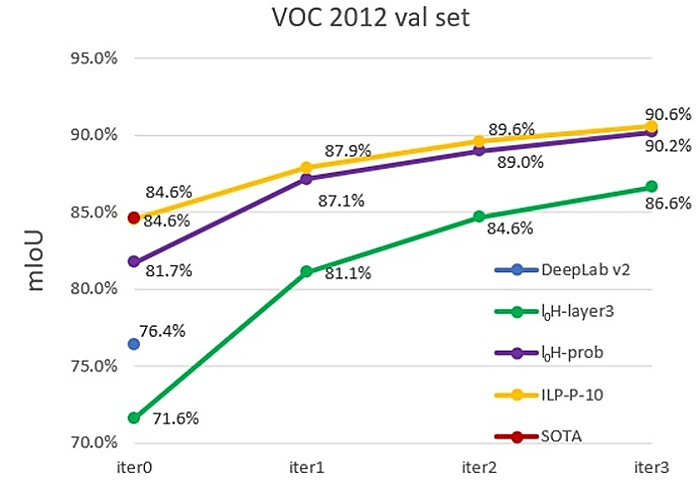}
\caption{Interactive segmentation on VOC 2012 \emph{val} set. The mIoU increases significantly with just one correction scribble per iteration. Blue dot is our baseline and red dot the current SOTA DCNN.}
\label{interactive_img}
\end{figure}

We simulate $3$ rounds of scribbles ($1$ scribble per image per round) to correct the largest error of previous segmentation.
We report their mIoU, compared to the baseline DCNN (DeepLab V$2$) and current reported state-of-the-art (SOTA) deep learning approach (DeepLab V$3+$ with $84.6\%$) \citep{2018arXiv180202611C} on VOC validation set. Results in Fig.~\ref{interactive_img} show that our algorithms benefit from additional scribbles and the performance increases significantly. The best result is reported by ILP-P-$10$ ($90.6\%$), a $6.0\%$ mIoU gain with just $3$ correction scribbles. See Fig.~\ref{scribbles} as an illustration of the procedure.

\subsubsection{Cityscapes}
We takes roughly 2 minutes to draw  modification scribbles, based on the previous segmentation results. Due to time constraint, we only pay attention to those largest segmentation errors through the whole image.
One can see in Table~\ref{modi-city} that all performance are better compared to the previous one, of which $\ell_0H$-prob-4000 reaches $80.5\%$ mIoU and $62.8\%$ PQ score. It is reported in~\cite{2018arXiv180100868K} that human consistency for very careful annotation only results in $69.7\%$ PQ. It is thus remarkable to reach $62.8\%$  PQ score under 15 minutes with just one round of scribble correction.

\begin{figure}[t]
\centering
\includegraphics[width=0.32\columnwidth]{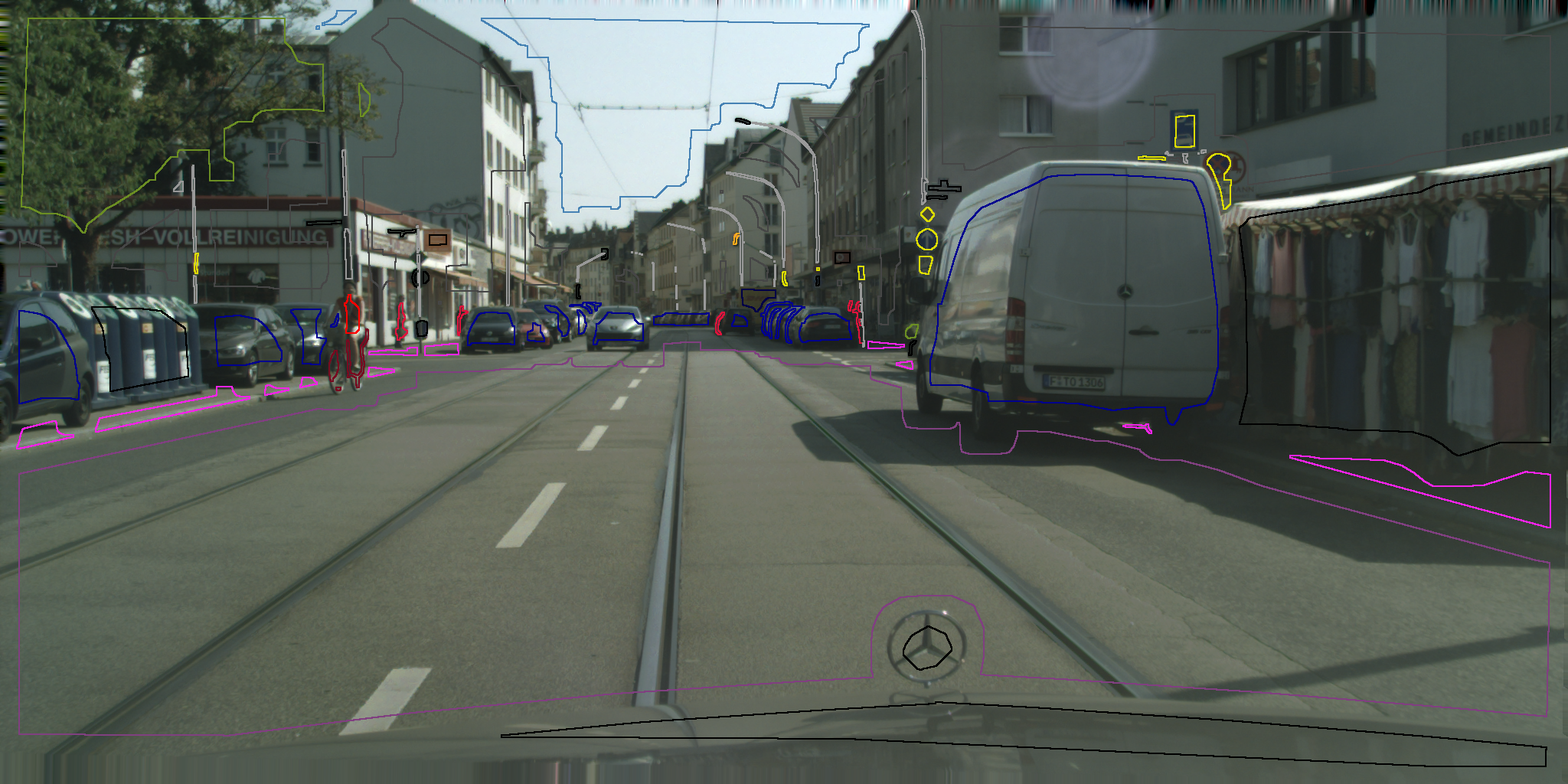}
\includegraphics[width=0.32\columnwidth]{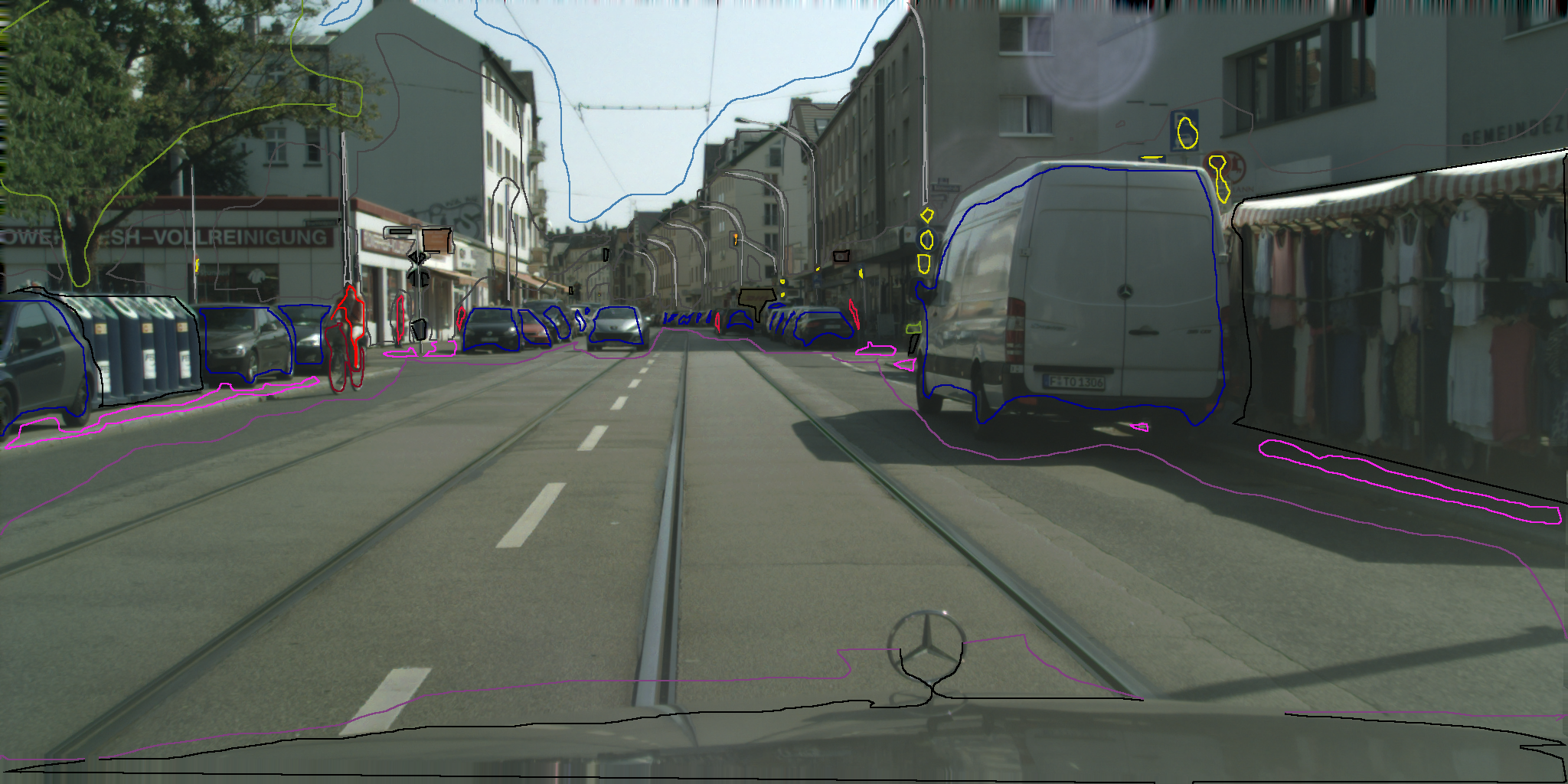}
\includegraphics[width=0.32\columnwidth]{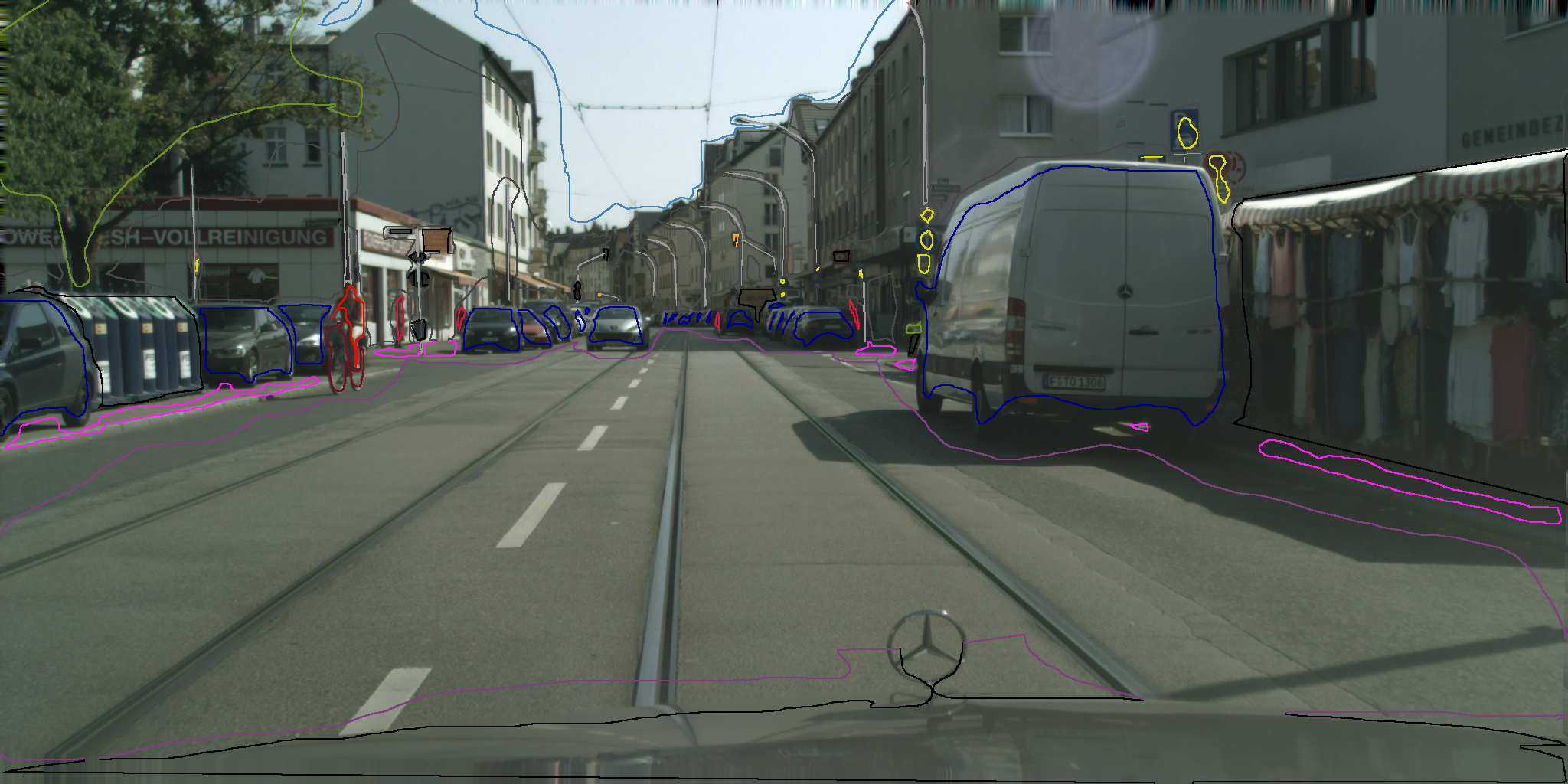}
\includegraphics[width=0.32\columnwidth]{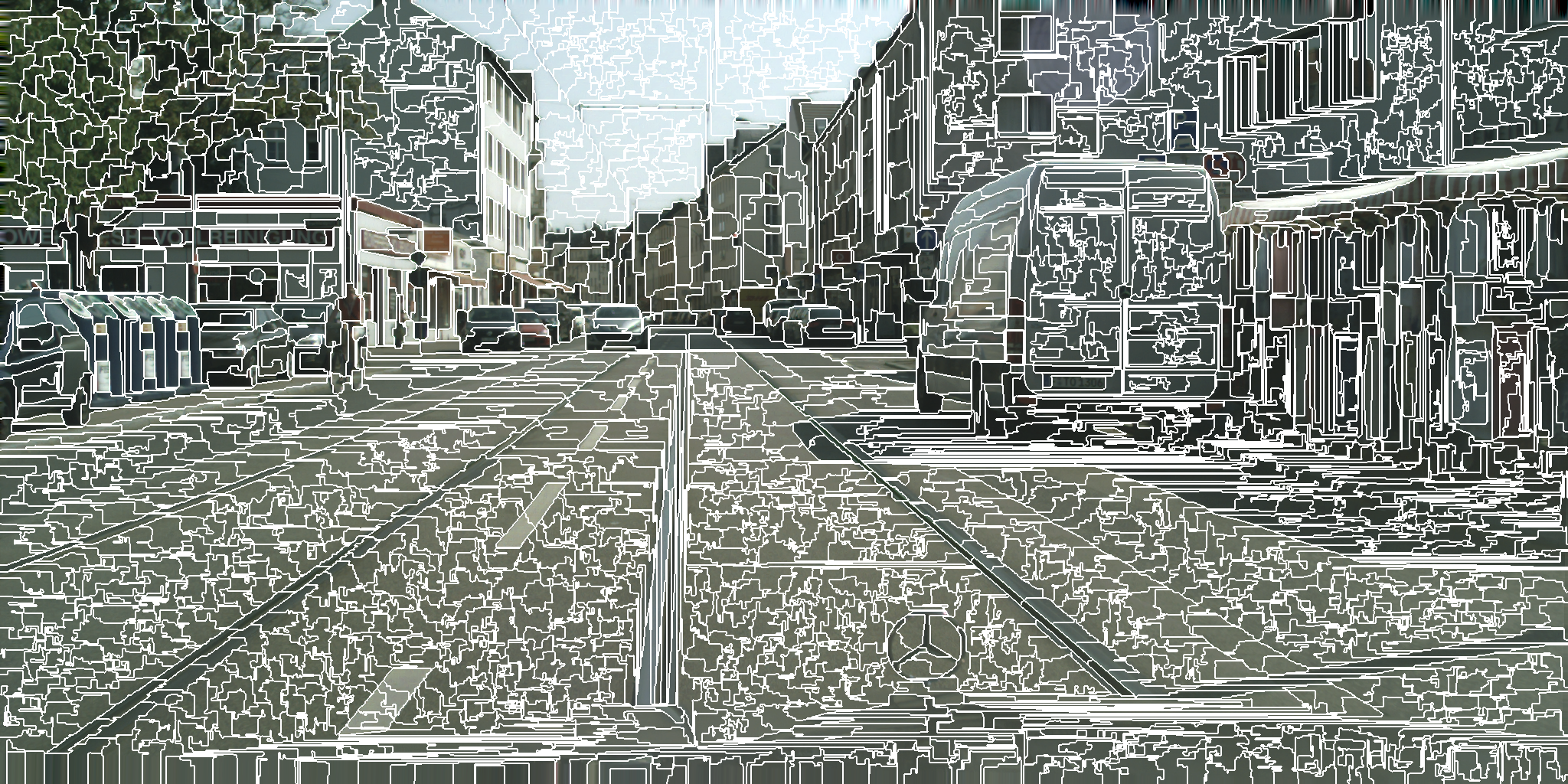}
\includegraphics[width=0.32\columnwidth]{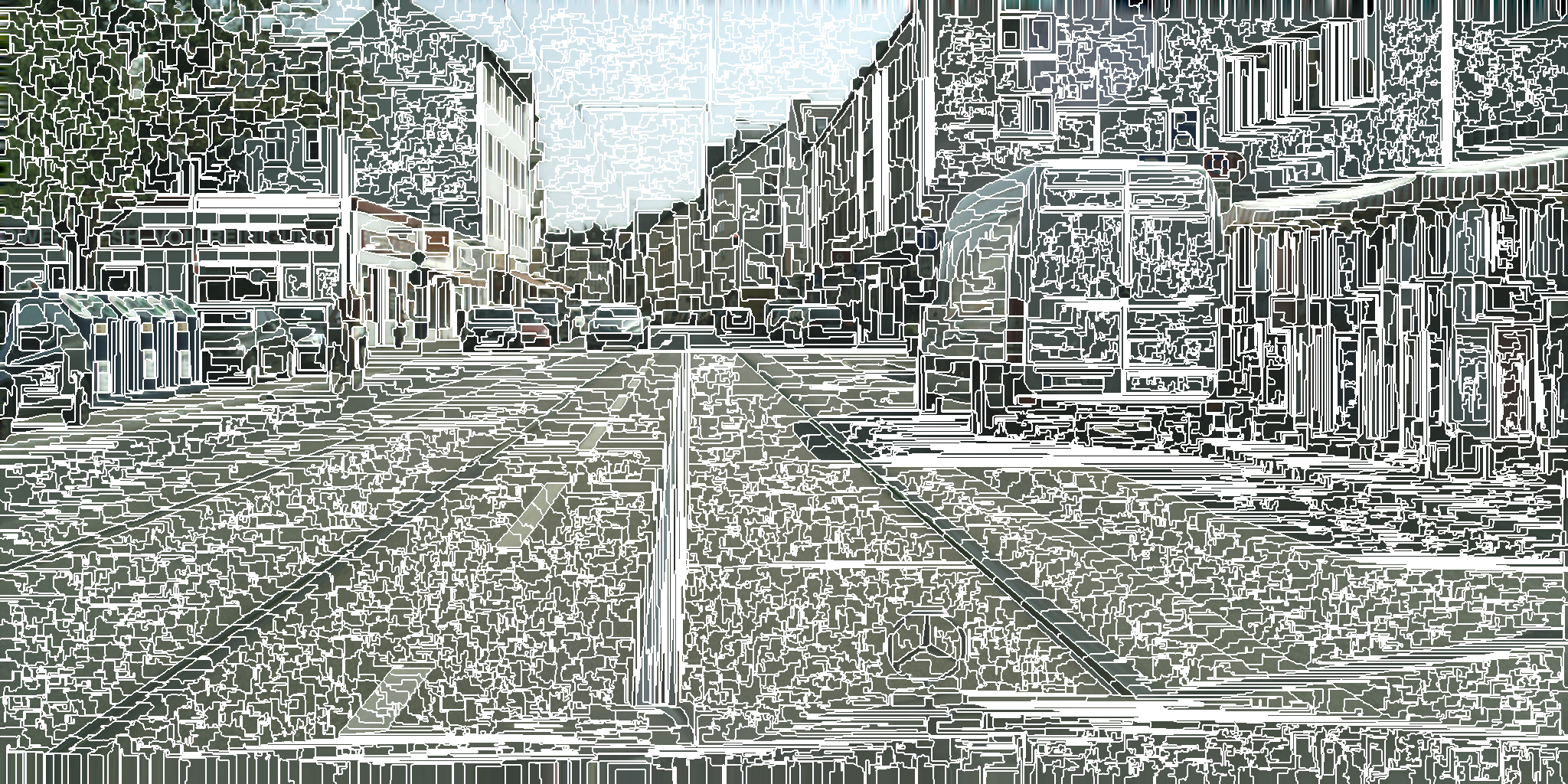}
\includegraphics[width=0.32\columnwidth]{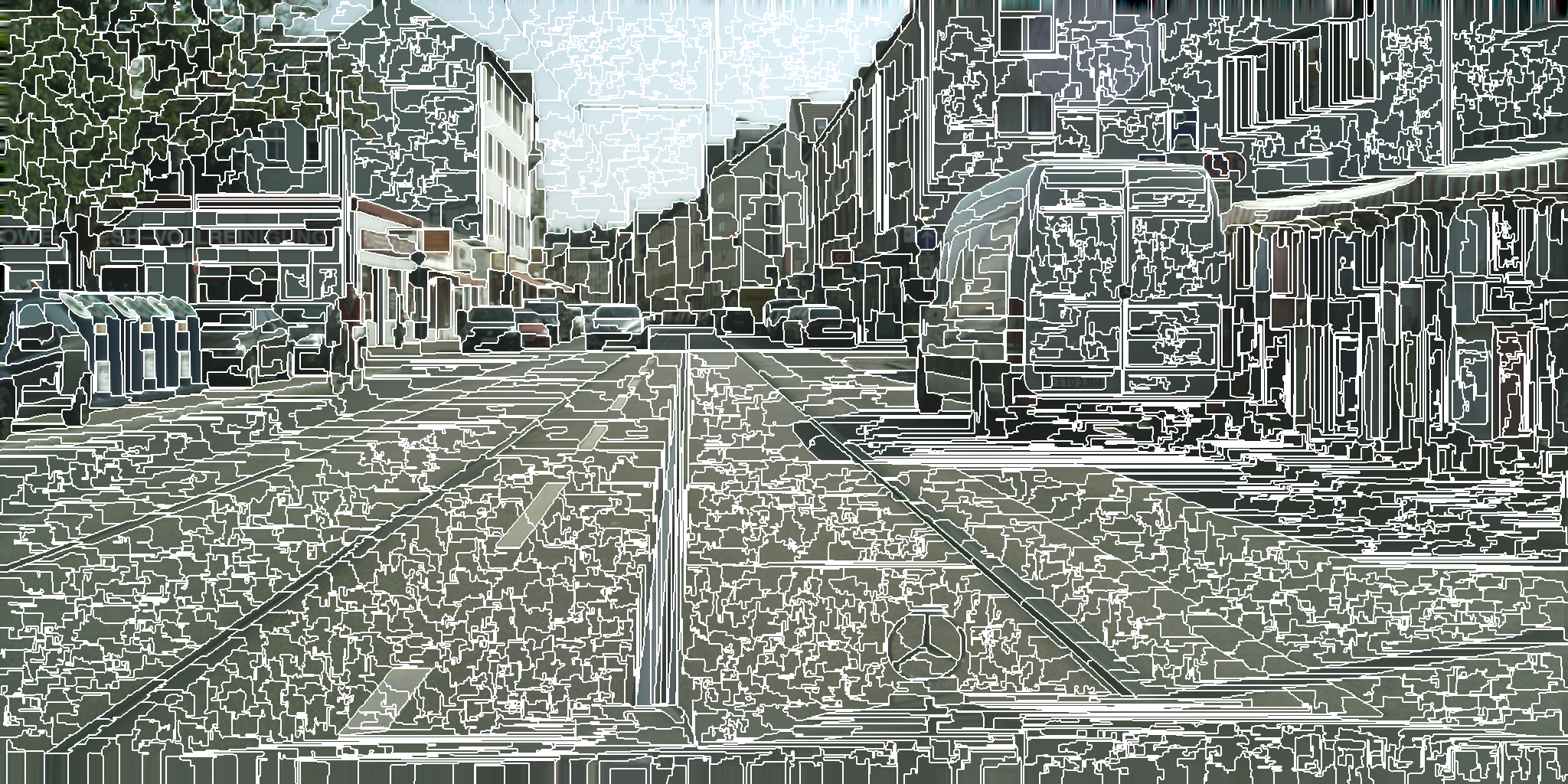}
\includegraphics[width=0.32\columnwidth]{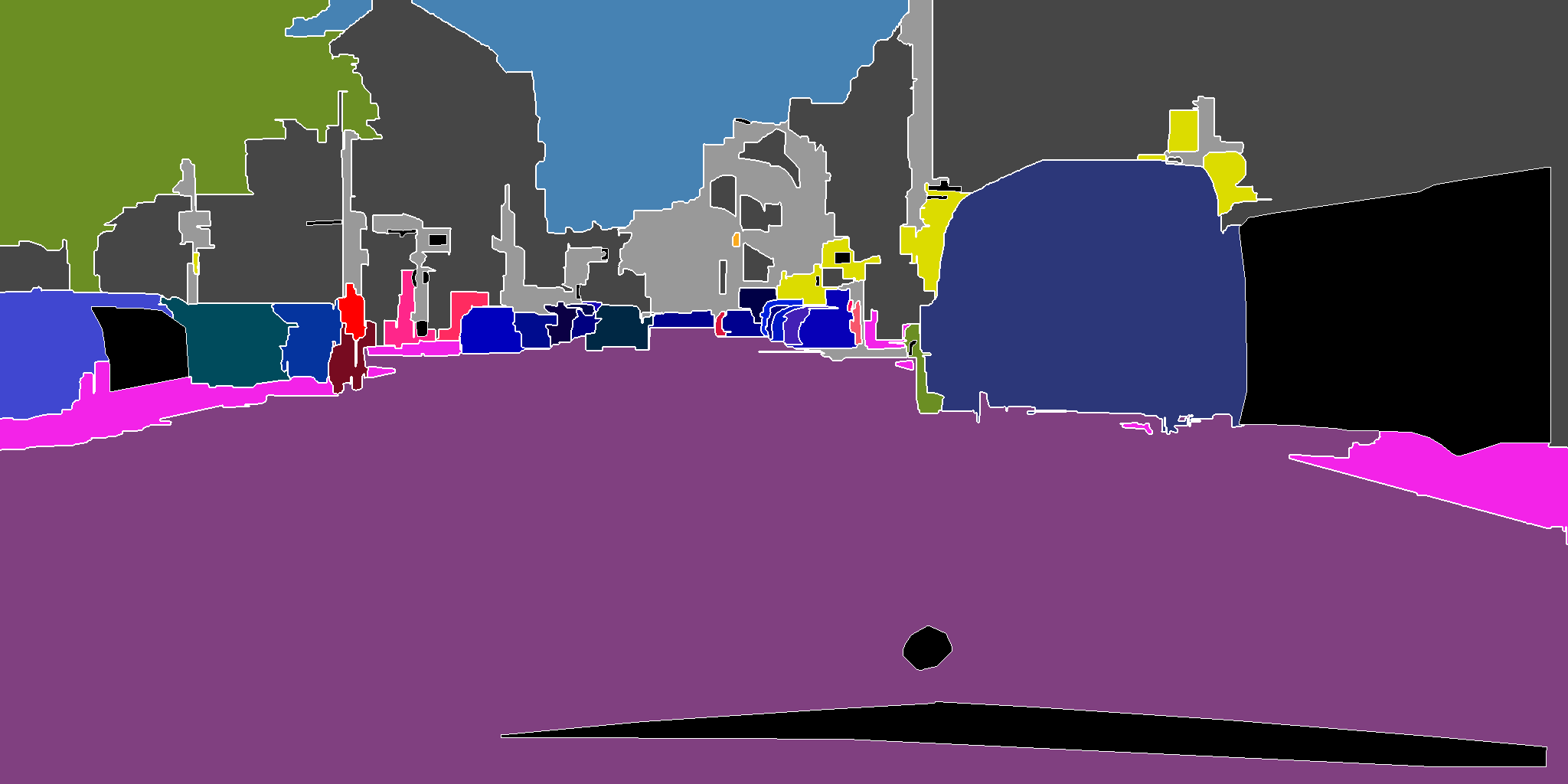}
\includegraphics[width=0.32\columnwidth]{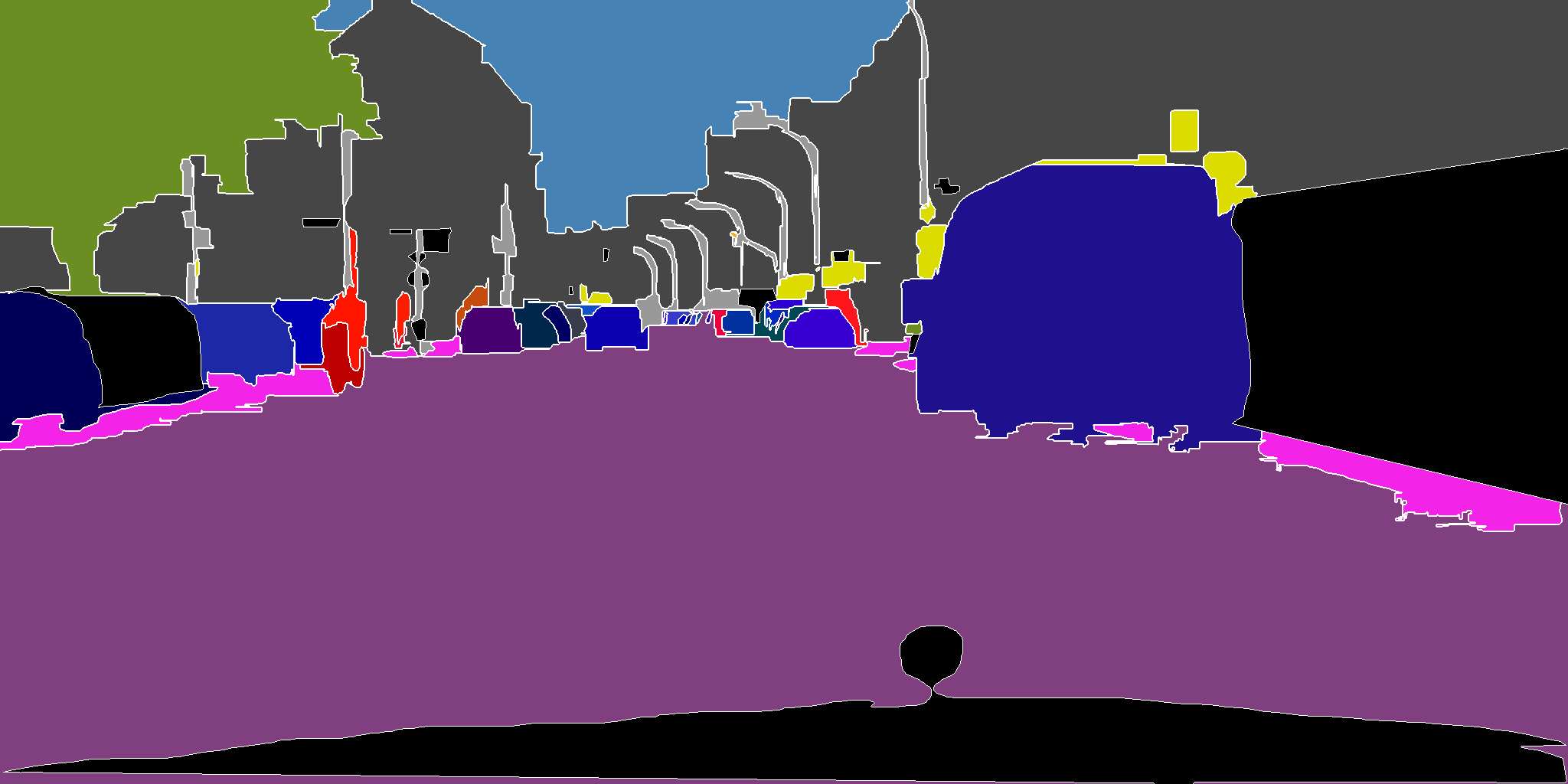}
\includegraphics[width=0.32\columnwidth]{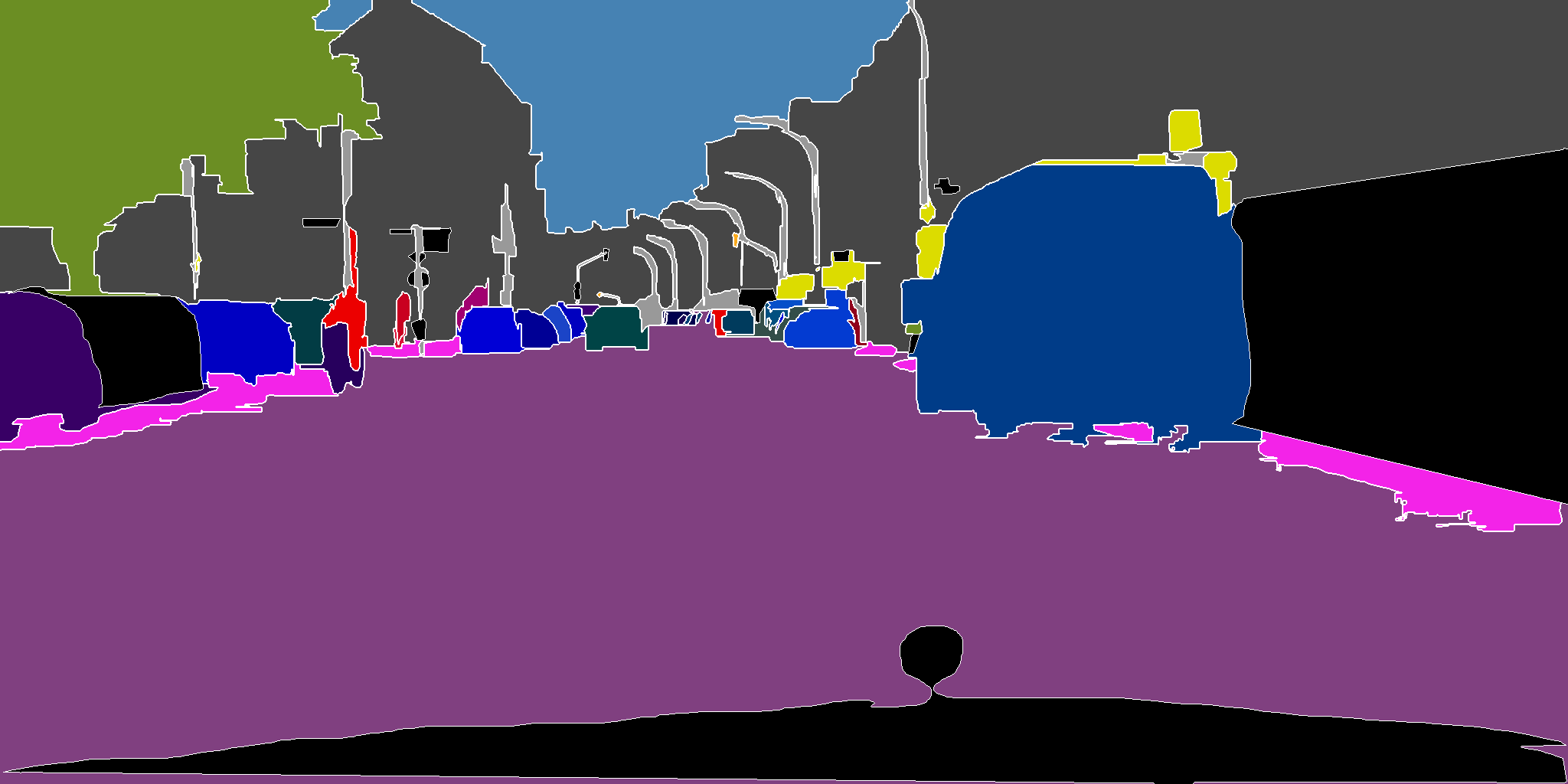}
\includegraphics[width=0.32\columnwidth]{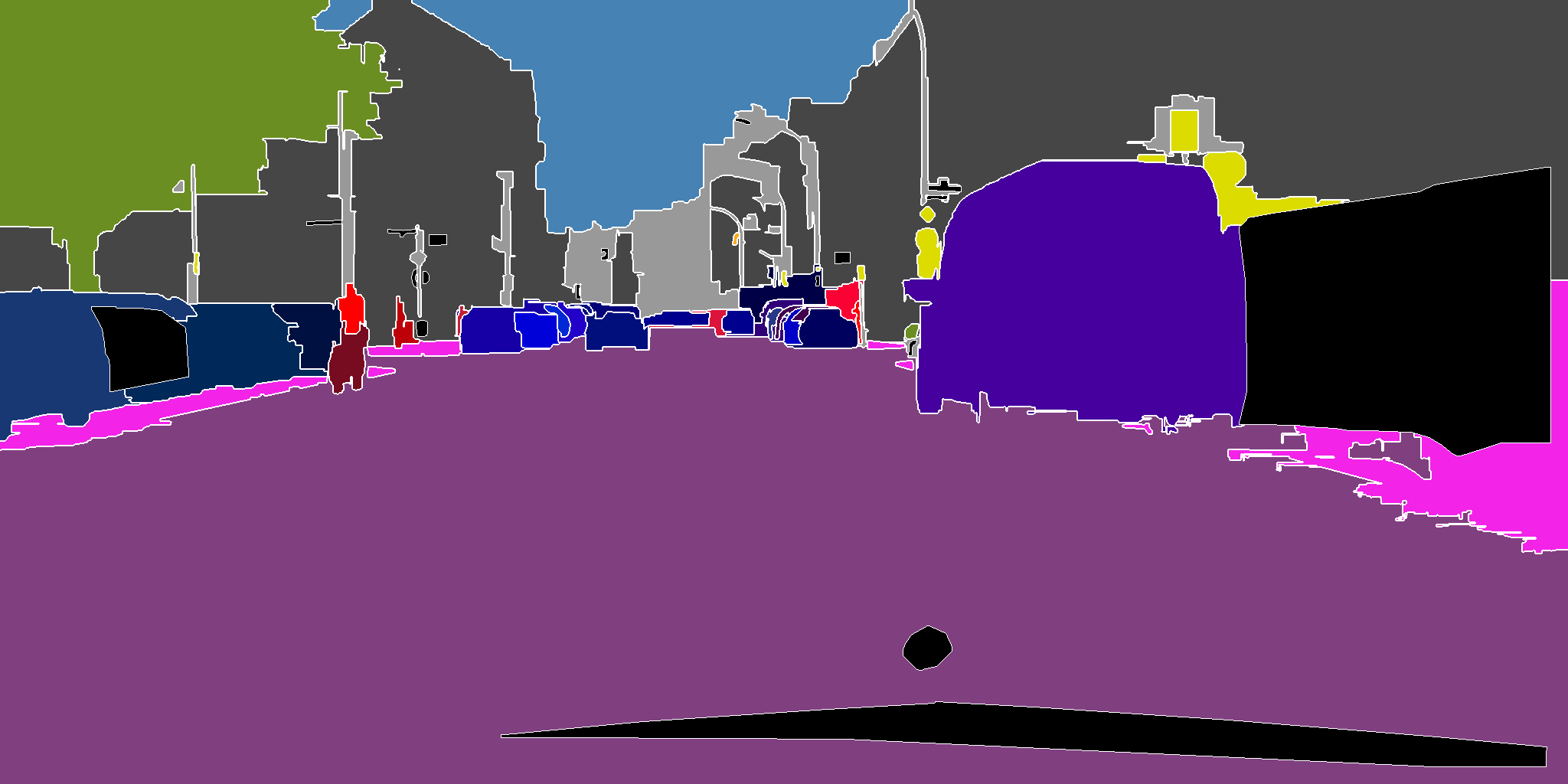}
\includegraphics[width=0.32\columnwidth]{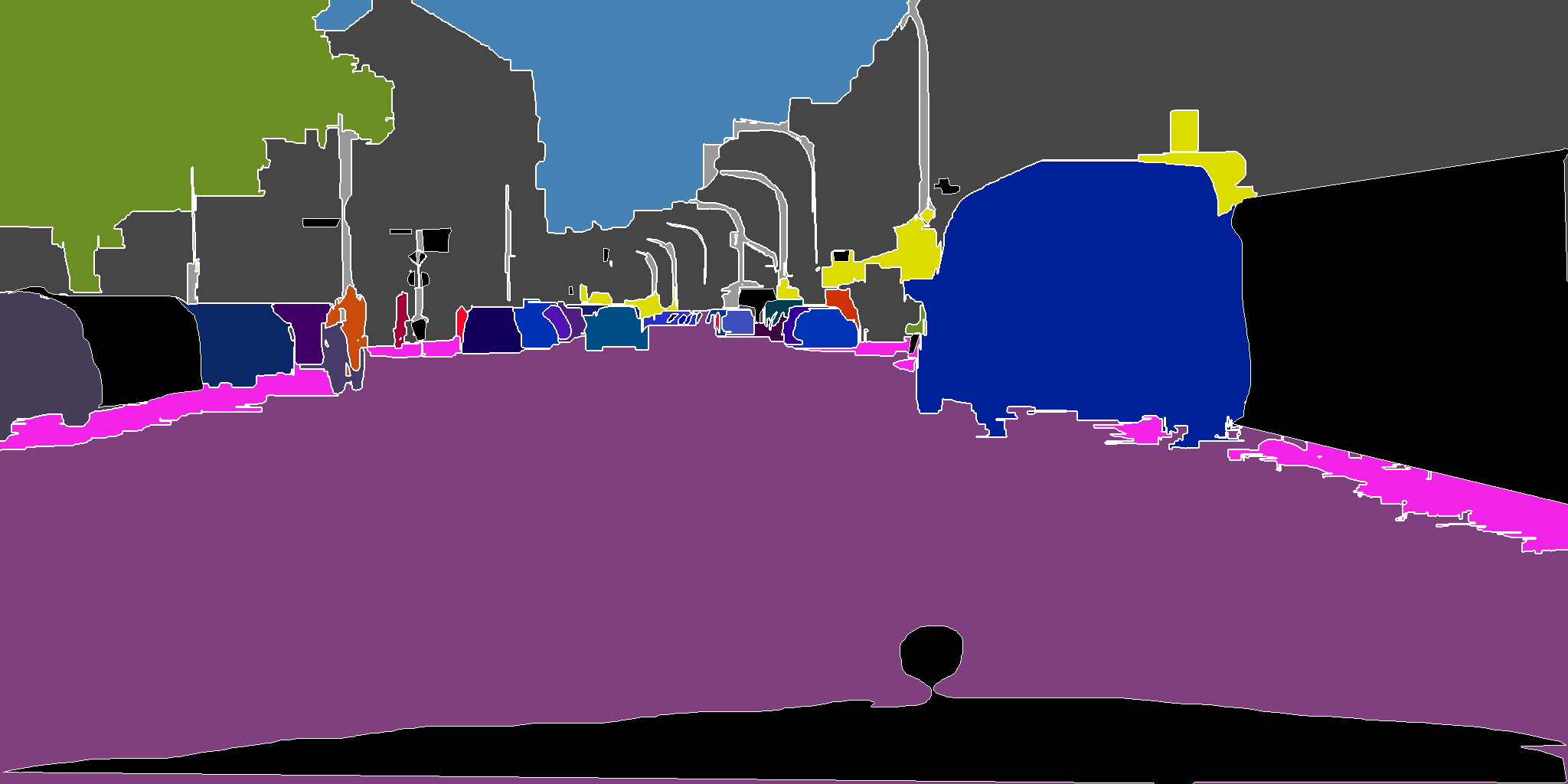}
\includegraphics[width=0.32\columnwidth]{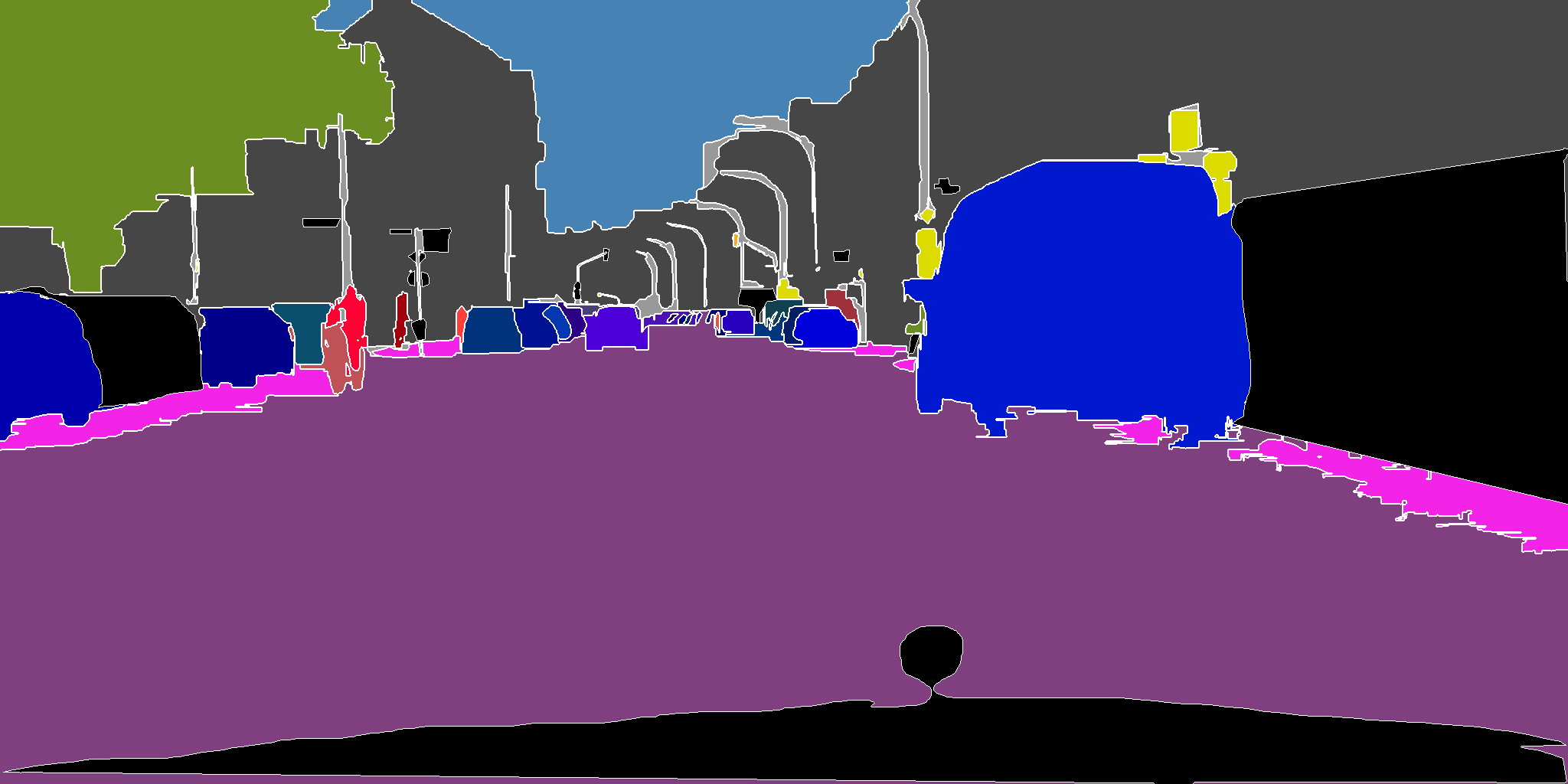}
\includegraphics[width=0.32\columnwidth]{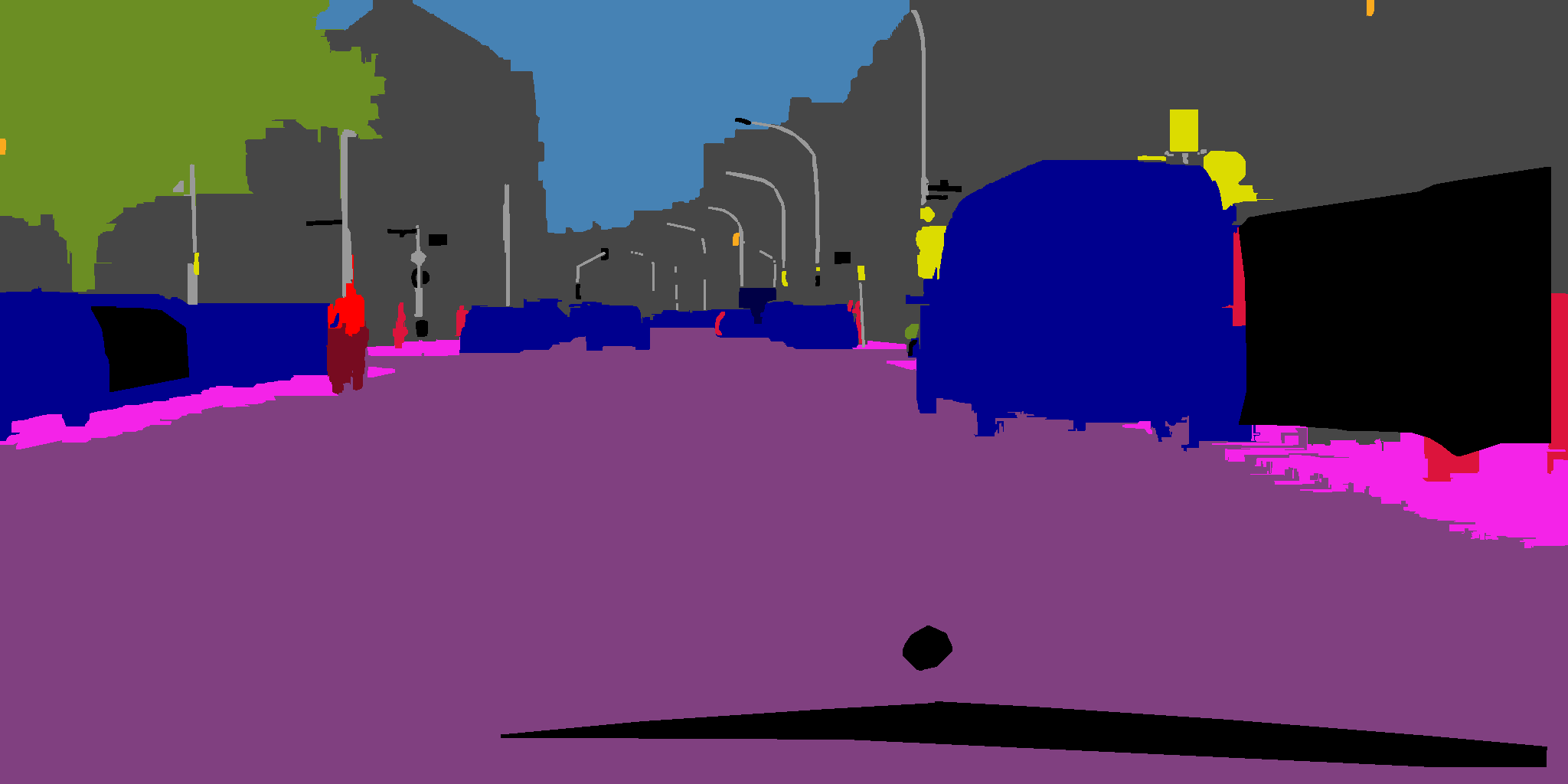}
\includegraphics[width=0.32\columnwidth]{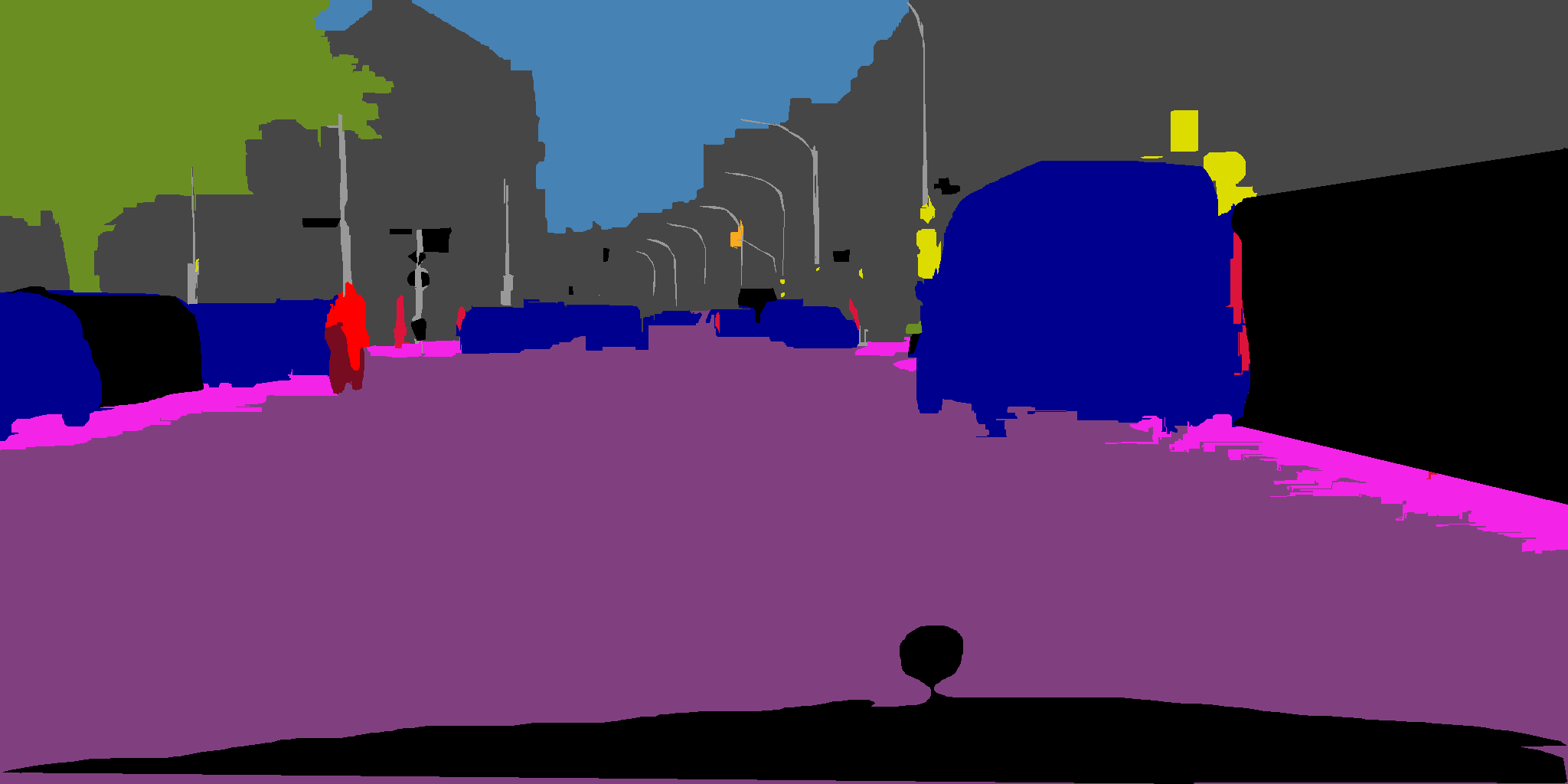}
\includegraphics[width=0.32\columnwidth]{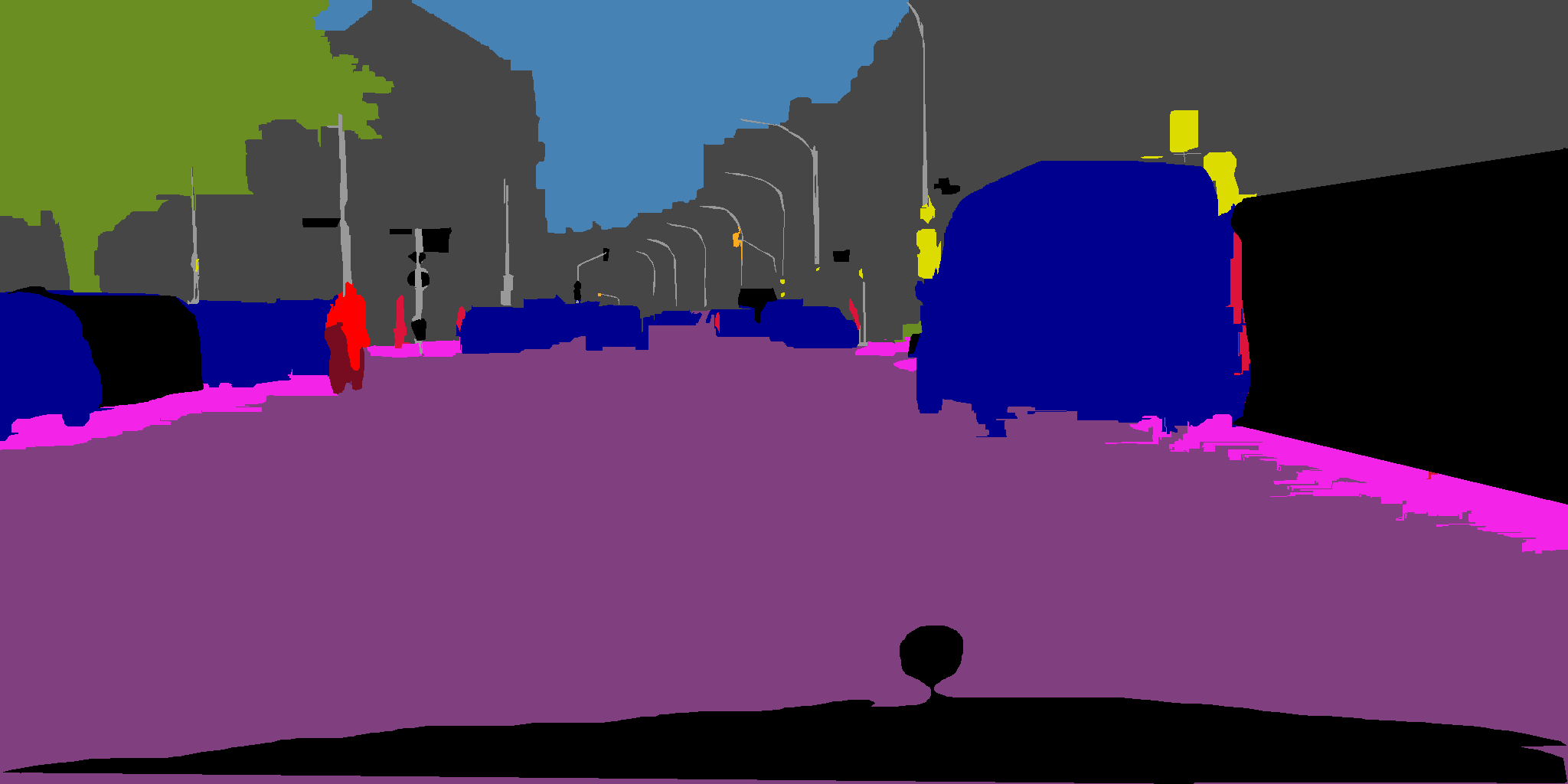}
\includegraphics[width=0.32\columnwidth]{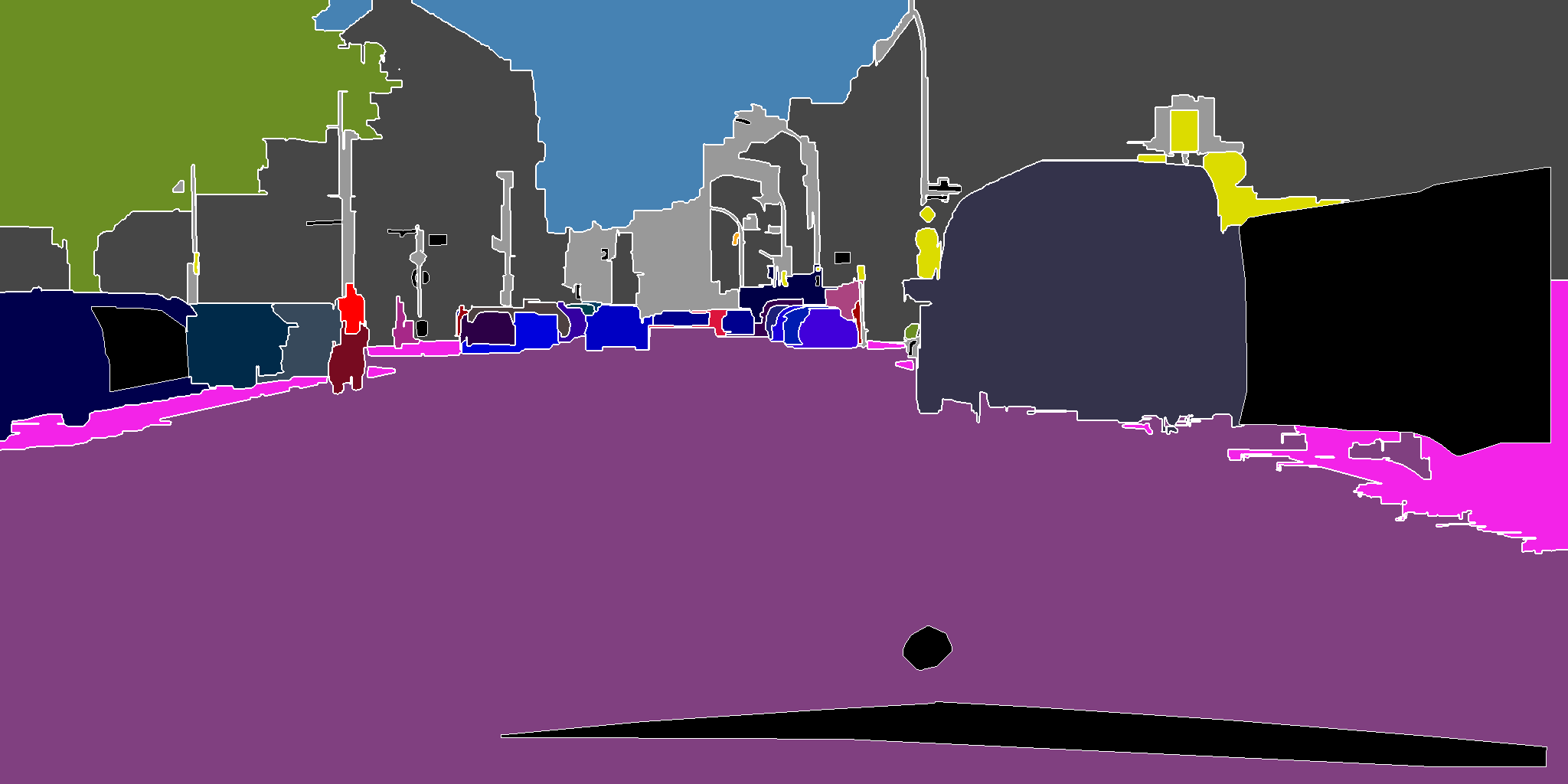}
\includegraphics[width=0.32\columnwidth]{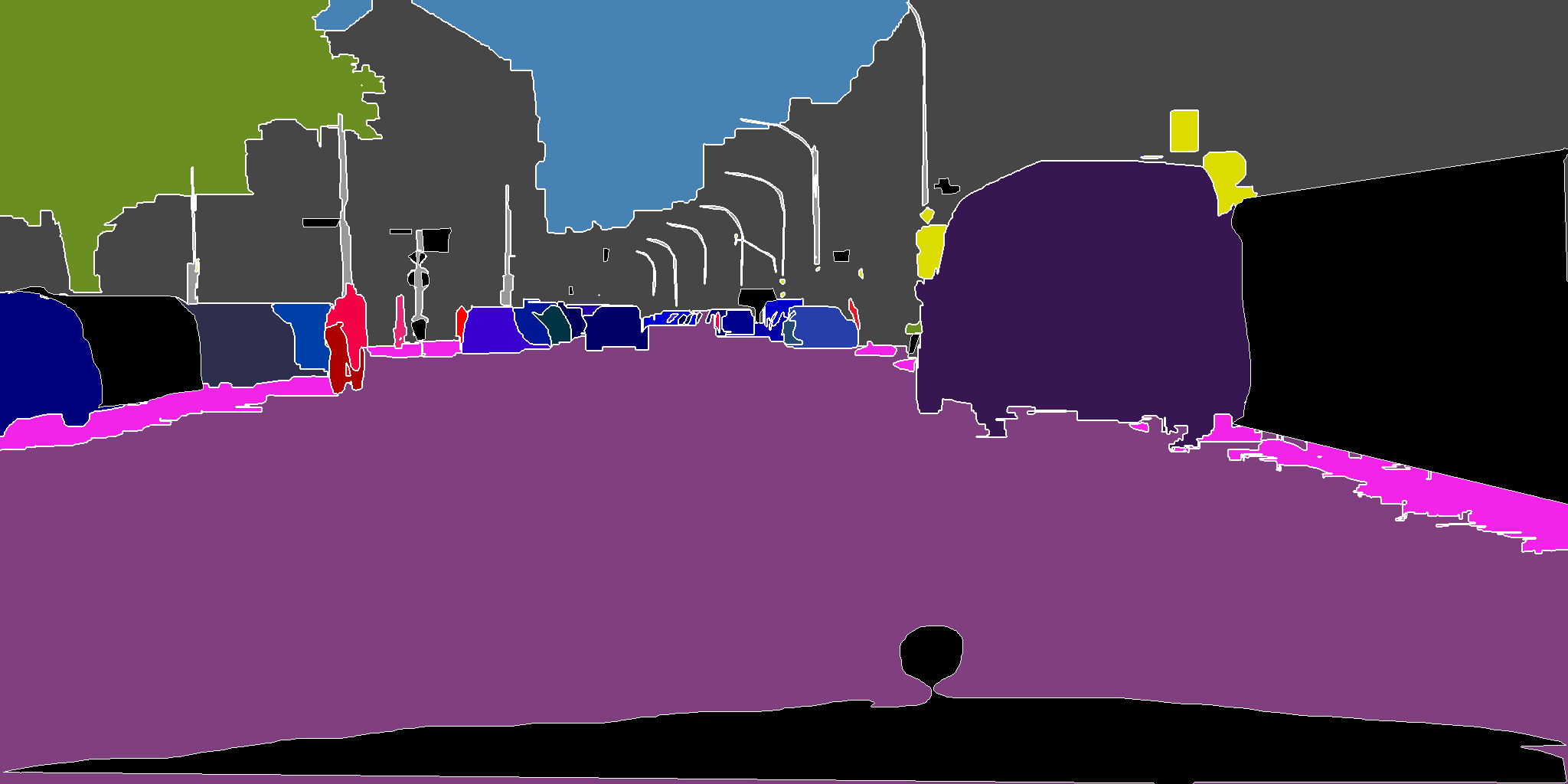}
\includegraphics[width=0.32\columnwidth]{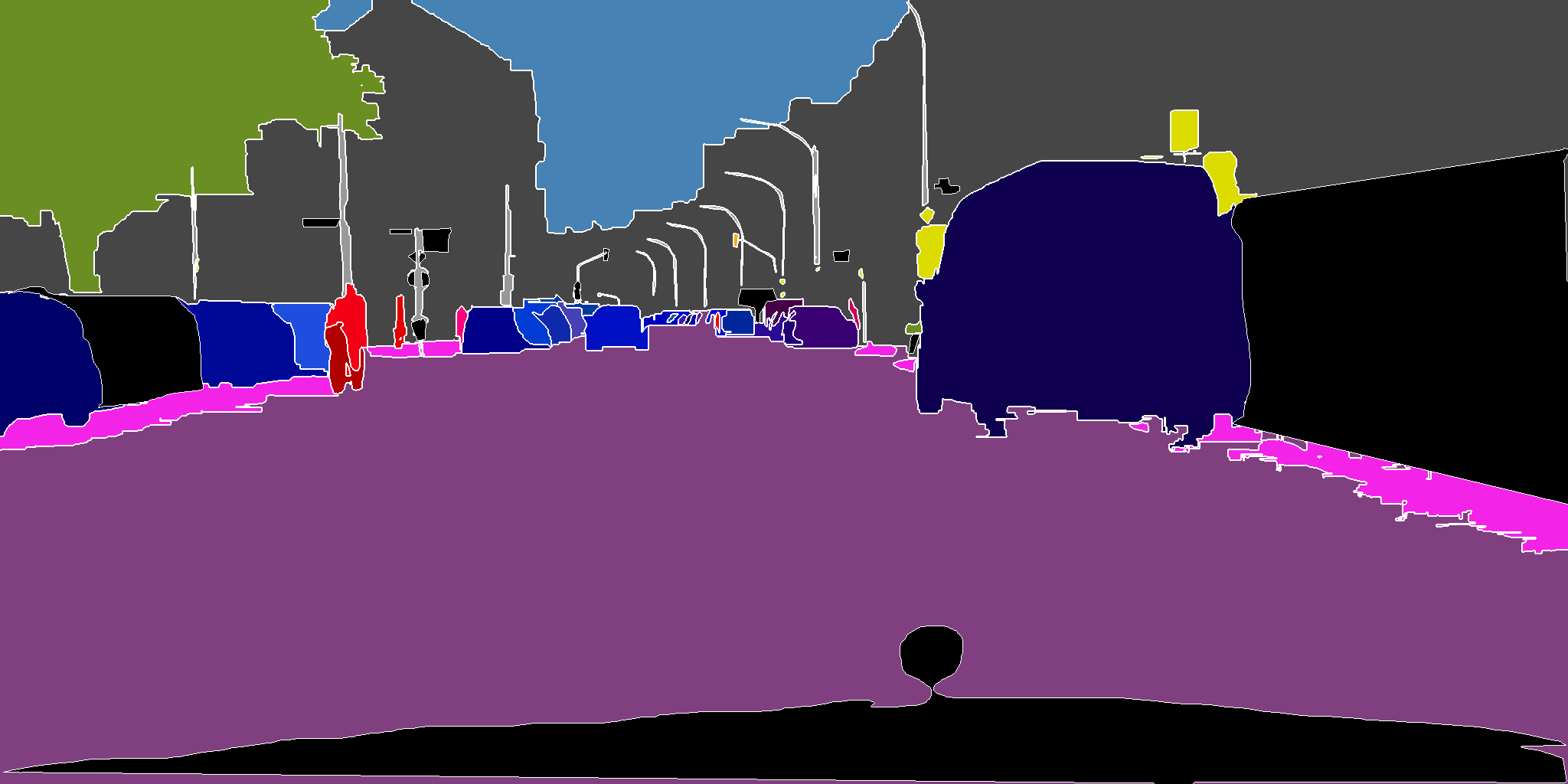}
\includegraphics[width=0.32\columnwidth]{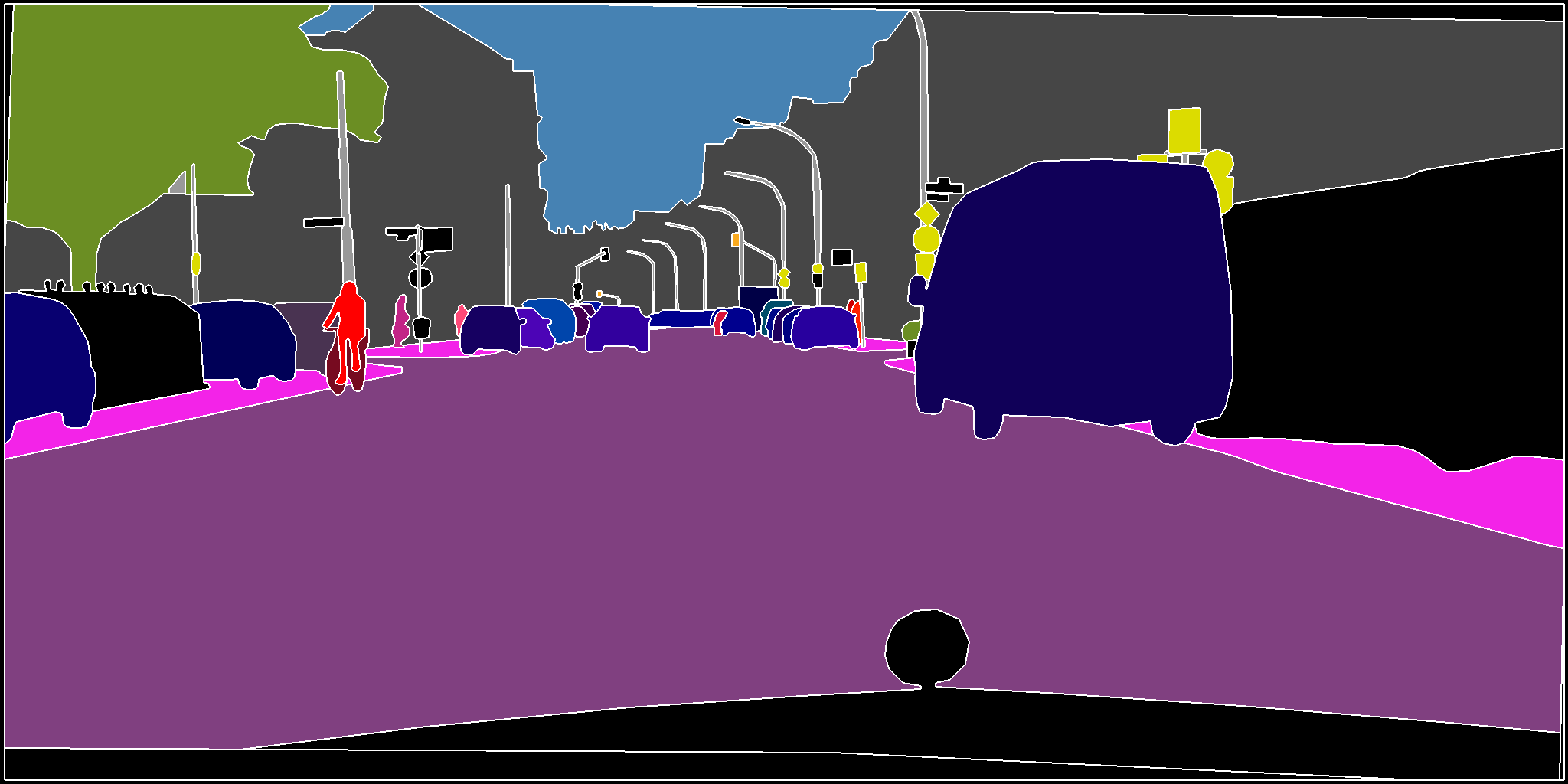}
\includegraphics[width=0.32\columnwidth]{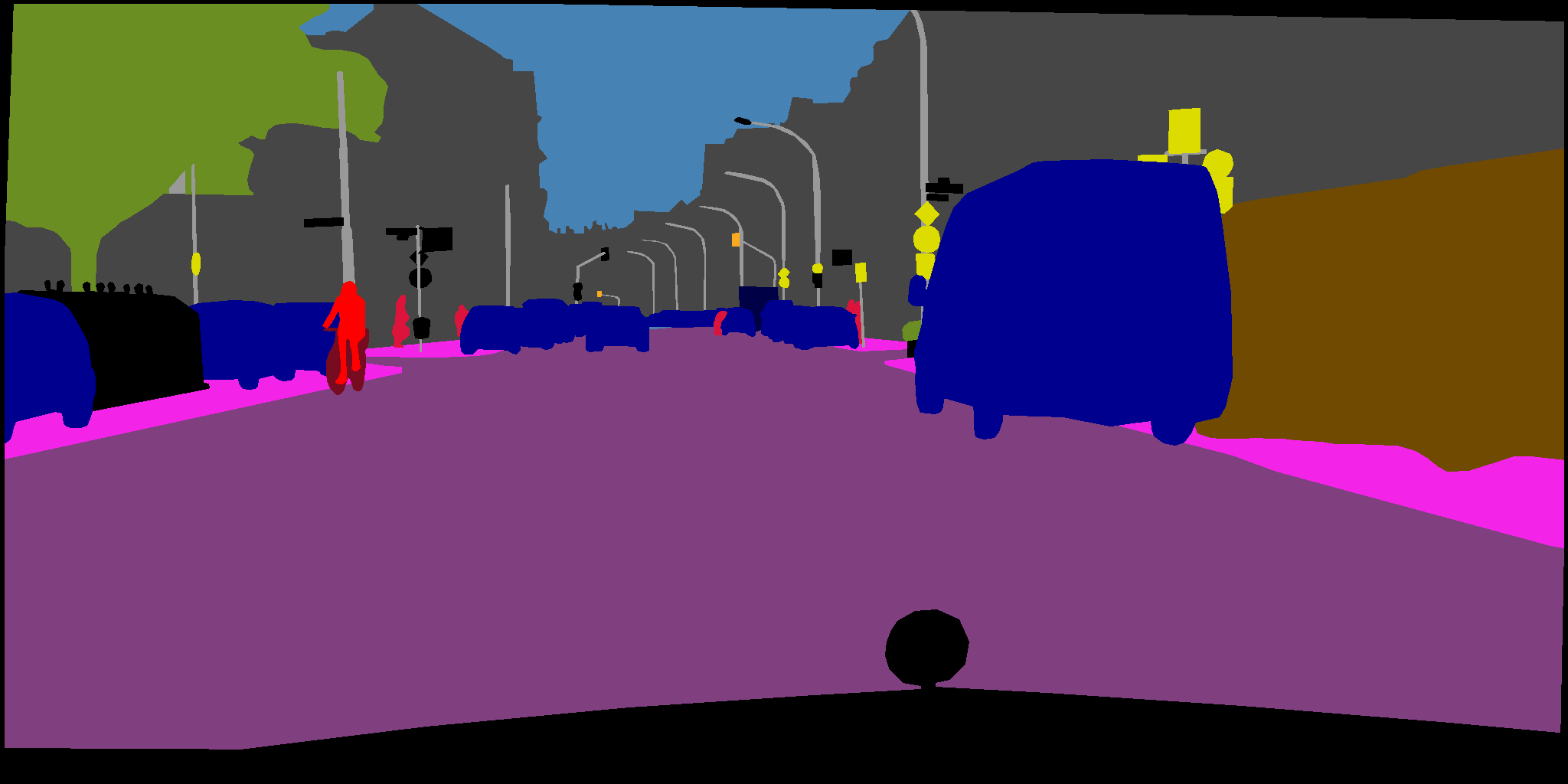}
\includegraphics[width=0.32\columnwidth]{img/f_10_1.png}
\caption{One cityscapes experiment with different setting and models. 1-3 columns: simulated, manual and (with) modification scribbles. 1-7 rows: cityscapes original image with scribbles, image with superpixels (4000 in the middle), $\ell_0H$-layer3, $\ell_0H$-prob, ILP-U, ILP-C and ground-truth. Note that ILP-U (fifth row) does not enforce connectivity, hence the existence of multiple ``person'' class (in red).}
\label{city_image}
\end{figure}

Among others that improves either mIoU or PQ more than $0.6\%$, ILP-U only benefit by a margin of $0.1\%$. 
This again indicates the importance of connectivity prior in a scribble based interactive framework.
Please also refer to the last two columns of Fig.~\ref{city_image}, to compare the impact of additional scribbles.

\section{Conclusions and future work}
We proposed two interactive graph based algorithms with global connectivity prior. With DCNN's features and initial scribbles as input, both achieve competitive semantic (and panoptic) segmentation results on VOC and Cityscapes validation set. The interactive framework further boost the performance with little overhead of correction scribbles.  It would be interesting to investigate our approach in a weakly supervised setting, where connectivity constraints are imposed on the outputs of a deep network to leverage unlabeled data. Training with such discrete high-order constraints can be explored via ADM (Alternating Direction Method) schemes, for instance, in way similar to \cite{DmitriCVPR19}, which showed promising performances of imposing discrete  on the outputs of convolutional networks.  
Finally, note that all the reported performance are upper bounded by the accuracy of superpixels. Hence, using better superpixel algorithms or  increasing its number  may further influence the performance. 


%






\ifCLASSOPTIONcaptionsoff
  \newpage
\fi



\bibliographystyle{IEEEtran}
\bibliography{research}
%

%











\end{document}

%% file: math_commands.tex
\usepackage{amsmath,amsfonts,bm}

\def\eqref#1{equation~\ref{#1}}

\def\1{\bm{1}}

\DeclareMathAlphabet{\mathsfit}{\encodingdefault}{\sfdefault}{m}{sl}
\SetMathAlphabet{\mathsfit}{bold}{\encodingdefault}{\sfdefault}{bx}{n}













%% file: main.bbl
\begin{thebibliography}{10}
\providecommand{\url}[1]{#1}
\csname url@samestyle\endcsname
\providecommand{\newblock}{\relax}
\providecommand{\bibinfo}[2]{#2}
\providecommand{\BIBentrySTDinterwordspacing}{\spaceskip=0pt\relax}
\providecommand{\BIBentryALTinterwordstretchfactor}{4}
\providecommand{\BIBentryALTinterwordspacing}{\spaceskip=\fontdimen2\font plus
\BIBentryALTinterwordstretchfactor\fontdimen3\font minus
  \fontdimen4\font\relax}
\providecommand{\BIBforeignlanguage}[2]{{%
\expandafter\ifx\csname l@#1\endcsname\relax
\typeout{** WARNING: IEEEtran.bst: No hyphenation pattern has been}%
\typeout{** loaded for the language `#1'. Using the pattern for}%
\typeout{** the default language instead.}%
\else
\language=\csname l@#1\endcsname
\fi
#2}}
\providecommand{\BIBdecl}{\relax}
\BIBdecl

\bibitem{7780459}
K.~He, X.~Zhang, S.~Ren, and J.~Sun, ``Deep residual learning for image
  recognition,'' in \emph{The IEEE Conference on Computer Vision and Pattern
  Recognition (CVPR)}, June 2016, pp. 770--778.

\bibitem{7485869}
S.~Ren, K.~He, R.~Girshick, and J.~Sun, ``Faster r-cnn: Towards real-time
  object detection with region proposal networks,'' \emph{The IEEE Transactions
  on Pattern Analysis and Machine Intelligence (TPAMI)}, vol.~39, no.~06, pp.
  1137--1149, Jun 2017.

\bibitem{Shelhamer2017}
E.~Shelhamer, J.~Long, and T.~Darrell, ``Fully convolutional networks for
  semantic segmentation,'' \emph{The IEEE Transactions on Pattern Analysis and
  Machine Intelligence (TPAMI)}, vol.~39, no.~4, pp. 640--651, Apr 2017.

\bibitem{7913730}
L.~Chen, G.~Papandreou, I.~Kokkinos, K.~Murphy, and A.~L. Yuille, ``Deeplab:
  Semantic image segmentation with deep convolutional nets, atrous convolution,
  and fully connected crfs,'' \emph{The IEEE Transactions on Pattern Analysis
  and Machine Intelligence (TPAMI)}, vol.~40, no.~4, pp. 834--848, April 2018.

\bibitem{zhao2017pspnet}
H.~Zhao, J.~Shi, X.~Qi, X.~Wang, and J.~Jia, ``Pyramid scene parsing network,''
  in \emph{The IEEE Conference on Computer Vision and Pattern Recognition
  (CVPR)}, 2017.

\bibitem{2018arXiv180100868K}
A.~{Kirillov}, K.~{He}, R.~{Girshick}, C.~{Rother}, and P.~{Doll{\'a}r},
  ``{Panoptic Segmentation},'' \emph{arXiv e-prints}, p. arXiv:1801.00868, Jan.
  2018.

\bibitem{2019arXiv190205093Y}
T.~Yang, M.~D. Collins, Y.~Zhu, J.~Hwang, T.~Liu, X.~Zhang, V.~Sze,
  G.~Papandreou, and L.~Chen, ``Deeperlab: Single-shot image parser,''
  \emph{arXiv e-prints}, Feb 2019.

\bibitem{2019arXiv190103784X}
Y.~Xiong, R.~Liao, H.~Zhao, R.~Hu, M.~Bai, E.~Yumer, and R.~Urtasun, ``Upsnet:
  A unified panoptic segmentation network,'' \emph{arXiv e-prints}, Jan 2019.

\bibitem{2019arXiv190102446K}
A.~Kirillov, R.~Girshick, K.~He, and P.~Doll{\'a}r, ``Panoptic feature pyramid
  networks,'' \emph{arXiv e-prints}, Jan 2019.

\bibitem{Chen2018MaskLabIS}
L.~Chen, A.~Hermans, G.~Papandreou, F.~Schroff, P.~Wang, and H.~Adam,
  ``Masklab: Instance segmentation by refining object detection with semantic
  and direction features,'' \emph{The IEEE Conference on Computer Vision and
  Pattern Recognition (CVPR)}, pp. 4013--4022, 2018.

\bibitem{Cordts2016Cityscapes}
M.~Cordts, M.~Omran, S.~Ramos, T.~Rehfeld, M.~Enzweiler, R.~Benenson,
  U.~Franke, S.~Roth, and B.~Schiele, ``The cityscapes dataset for semantic
  urban scene understanding,'' in \emph{The IEEE Conference on Computer Vision
  and Pattern Recognition (CVPR)}, 2016.

\bibitem{969114}
Y.~Boykov, O.~Veksler, and R.~Zabih, ``Fast approximate energy minimization via
  graph cuts,'' \emph{The IEEE Transactions on Pattern Analysis and Machine
  Intelligence (TPAMI)}, vol.~23, no.~11, pp. 1222--1239, Nov 2001.

\bibitem{Yedidia:2003}
J.~S. Yedidia, W.~T. Freeman, and Y.~Weiss, ``Exploring artificial intelligence
  in the new millennium,'' pp. 239--269, 2003.

\bibitem{potts_1952}
R.~B. Potts, ``Some generalized order-disorder transformations,''
  \emph{Mathematical Proceedings of the Cambridge Philosophical Society},
  vol.~48, no.~1, pp. 106–--109, 1952.

\bibitem{698673}
Y.~Boykov, O.~Veksler, and R.~Zabih, ``Markov random fields with efficient
  approximations,'' in \emph{The IEEE Conference on Computer Vision and Pattern
  Recognition (CVPR)}, June 1998, pp. 648--655.

\bibitem{ILSVRC15}
O.~Russakovsky, J.~Deng, H.~Su, J.~Krause, S.~Satheesh, S.~Ma, Z.~Huang,
  A.~Karpathy, A.~Khosla, M.~Bernstein, A.~C. Berg, and F.~Li, ``Imagenet large
  scale visual recognition challenge,'' \emph{International Journal of Computer
  Vision}, vol. 115, no.~3, pp. 211--252, 2015.

\bibitem{10.100748}
T.~Lin, M.~Maire, S.~Belongie, J.~Hays, P.~Perona, D.~Ramanan, P.~Doll{\'a}r,
  and C.~L. Zitnick, ``Microsoft coco: Common objects in context,'' in
  \emph{The European Conference on Computer Vision (ECCV)}, 2014, pp. 740--755.

\bibitem{Everingham15}
M.~Everingham, S.~M.~A. Eslami, L.~V. Gool, C.~K.~I. Williams, J.~Winn, and
  A.~Zisserman, ``The pascal visual object classes challenge: A
  retrospective,'' \emph{International Journal of Computer Vision}, vol. 111,
  no.~1, pp. 98--136, Jan 2015.

\bibitem{Lin2016ScribbleSupSC}
D.~Lin, J.~Dai, J.~Jia, K.~He, and J.~Sun, ``Scribblesup: Scribble-supervised
  convolutional networks for semantic segmentation,'' \emph{The IEEE Conference
  on Computer Vision and Pattern Recognition (CVPR)}, pp. 3159--3167, 2016.

\bibitem{8215750}
X.~Chen and T.~Wang, ``Combining active learning and semi-supervised learning
  by using selective label spreading,'' in \emph{2017 IEEE International
  Conference on Data Mining Workshops (ICDMW)}, 2017, pp. 850--857.

\bibitem{Rother:2004}
C.~Rother, V.~Kolmogorov, and A.~Blake, ``"grabcut": Interactive foreground
  extraction using iterated graph cuts,'' \emph{ACM Trans. Graph.}, pp.
  309--314, Aug 2004.

\bibitem{Man+18}
K.-K. Maninis, S.~Caelles, J.~Pont-Tuset, and L.~V. Gool, ``Deep extreme cut:
  From extreme points to object segmentation,'' in \emph{The IEEE Conference on
  Computer Vision and Pattern Recognition (CVPR)}, 2018.

\bibitem{extremeclick}
D.~Papadopoulos, J.~Uijlings, F.~Keller, and V.~Ferrari, ``Extreme clicking for
  efficient object annotation,'' in \emph{The IEEE International Conference on
  Computer Vision (ICCV)}, Dec. 2017, pp. 4940--4949.

\bibitem{Russell2008}
B.~C. Russell, A.~Torralba, K.~P. Murphy, and W.~T. Freeman, ``Labelme: A
  database and web-based tool for image annotation,'' \emph{International
  Journal of Computer Vision}, vol.~77, no.~1, pp. 157--173, May 2008.

\bibitem{937505}
Y.~Y. Boykov and M.~. Jolly, ``Interactive graph cuts for optimal boundary amp;
  region segmentation of objects in n-d images,'' in \emph{The Eighth IEEE
  International Conference on Computer Vision (ICCV)}, July 2001, pp. 105--112
  vol.1.

\bibitem{4359322}
A.~Levin, D.~Lischinski, and Y.~Weiss, ``A closed-form solution to natural
  image matting,'' \emph{The IEEE Transactions on Pattern Analysis and Machine
  Intelligence (TPAMI)}, vol.~30, no.~2, pp. 228--242, Feb 2008.

\bibitem{acuna2018efficient}
D.~Acuna, H.~Ling, A.~Kar, and S.~Fidler, ``Efficient interactive annotation of
  segmentation datasets with polygon-rnn++,'' \emph{The IEEE Conference on
  Computer Vision and Pattern Recognition (CVPR)}, 2018.

\bibitem{2019arXiv190306874L}
H.~Ling, J.~Gao, A.~Kar, W.~Chen, and S.~Fidler, ``{Fast Interactive Object
  Annotation with Curve-GCN},'' \emph{arXiv e-prints}, Mar 2019.

\bibitem{agustsson2019interactive}
E.~Agustsson, J.~R. Uijlings, and V.~Ferrari, ``Interactive full image
  segmentation by considering all regions jointly,'' in \emph{Proceedings of
  the IEEE Conference on Computer Vision and Pattern Recognition}, 2019, pp.
  11\,622--11\,631.

\bibitem{Zhou2009}
Z.~Zhou, ``Ensemble learning,'' \emph{Encyclopedia of Biometrics}, pp.
  270--273, 2009.

\bibitem{Dai2015BoxSupEB}
J.~Dai, K.~He, and J.~Sun, ``Boxsup: Exploiting bounding boxes to supervise
  convolutional networks for semantic segmentation,'' \emph{The IEEE
  International Conference on Computer Vision (ICCV)}, pp. 1635--1643, 2015.

\bibitem{khoreva_CVPR17}
A.~Khoreva, R.~Benenson, J.~Hosang, M.~Hein, and B.~Schiele, ``Simple does it:
  Weakly supervised instance and semantic segmentation,'' in \emph{The IEEE
  Conference on Computer Vision and Pattern Recognition (CVPR)}, 2017.

\bibitem{Li_2018_ECCV}
Q.~Li, A.~Arnab, and P.~H. Torr, ``Weakly- and semi-supervised panoptic
  segmentation,'' in \emph{The European Conference on Computer Vision (ECCV)},
  Sept 2018.

\bibitem{20.500}
Y.~B. Can, K.~Chaitanya, B.~Mustafa, L.~M. Koch, E.~Konukoglu, and
  C.~Baumgartner, ``Learning to segment medical images with
  scribble-supervision alone,'' in \emph{Deep Learning in Medical Image
  Analysis and Multimodal Learning for Clinical Decision Support}, vol. 11045,
  2018, pp. 236--244.

\bibitem{DBLP-1803-07351}
R.~Shen, X.~Chen, X.~Zheng, and G.~Reinelt, ``Discrete potts model for
  generating superpixels on noisy images,'' \emph{CoRR}, vol. abs/1803.07351,
  2018.

\bibitem{7410389}
R.~M.~H. Nguyen and M.~S. Brown, ``Fast and effective l0 gradient minimization
  by region fusion,'' in \emph{The IEEE International Conference on Computer
  Vision (ICCV)}, Dec 2015, pp. 208--216.

\bibitem{rshen_ilp2017}
R.~Shen, B.~Tang, A.~Lodi, A.~Tramontani, and I.~B. Ayed, ``An ilp model for
  multi-label mrfs with connectivity constraints,'' \emph{IEEE Transactions on
  Image Processing}, vol.~29, pp. 6909--6917, 2020.

\bibitem{Boykov2006}
Y.~Boykov and O.~Veksler, ``Graph cuts in vision and graphics: Theories and
  applications,'' \emph{Handbook of Mathematical Models in Computer Vision},
  pp. 79--96, 2006.

\bibitem{NIPS2011_4296}
P.~Kr\"{a}henb\"{u}hl and V.~Koltun, ``Efficient inference in fully connected
  crfs with gaussian edge potentials,'' \emph{Advances in Neural Information
  Processing Systems}, pp. 109--117, 2011.

\bibitem{Rempfler2016TheMC}
M.~Rempfler, B.~Andres, and B.~H. Menze, ``The minimum cost connected subgraph
  problem in medical image analysis,'' in \emph{The International Conference on
  Medical Image Computing and Computer Assisted Intervention (MICCAI)}, 2016.

\bibitem{5828}
S.~Nowozin and C.~H. Lampert, ``Global connectivity potentials for random field
  models,'' in \emph{The IEEE Conference on Computer Vision and Pattern
  Recognition (CVPR)}, Jun 2009, pp. 818--825.

\bibitem{4587440}
S.~Vicente, V.~Kolmogorov, and C.~Rother, ``Graph cut based image segmentation
  with connectivity priors,'' in \emph{The IEEE Conference on Computer Vision
  and Pattern Recognition (CVPR)}, June 2008, pp. 1--8.

\bibitem{doi:10.1080}
R.~Anand, D.~Aggarwal, and V.~Kumar, ``A comparative analysis of optimization
  solvers,'' \emph{Journal of Statistics and Management Systems}, vol.~20,
  no.~4, pp. 623--635, 2017.

\bibitem{rshen2018}
R.~Shen, ``Milp formulations for unsupervised and interactive image
  segmentation and denoising,'' Ph.D. dissertation, Heidelberg University,
  2018.

\bibitem{Land60anautomatic}
A.~H. Land and A.~G. Doig, ``An automatic method for solving discrete
  programming problems,'' \emph{ECONOMETRICA}, vol.~28, no.~3, pp. 497--520,
  1960.

\bibitem{doi:10.1137}
J.~E. Kelley, ``The cutting-plane method for solving convex programs,''
  \emph{Journal of the Society for Industrial and Applied Mathematics}, vol.~8,
  no.~4, pp. 703--712, 1960.

\bibitem{STUTZ2017}
D.~Stutz, A.~Hermans, and B.~Leibe, ``Superpixels: an evaluation of the
  state-of-the-art,'' \emph{Computer Vision and Image Understanding}, pp.
  1--32, 2017.

\bibitem{SLIC}
R.~Achanta, A.~Shaji, K.~Smith, A.~Lucchi, P.~Fua, and S.~Susstrunk, ``Slic
  superpixels compared to state-of-the-art superpixel methods,'' \emph{The IEEE
  Transactions on Pattern Analysis and Machine Intelligence (TPAMI)}, vol.~34,
  no.~11, pp. 2274--2282, 2012.

\bibitem{seeds}
M.~V. den Bergh, X.~Boix, G.~Roig, B.~de~Capitani, and L.~V. Gool, ``Seeds:
  Superpixels extracted via energy-driven sampling,'' \emph{The European
  Conference on Computer Vision (ECCV)}, pp. 13--26, 2012.

\bibitem{Tu-CVPR-2018}
W.~Tu, M.~Liu, V.~Jampani, D.~Sun, S.~Chien, M.~Yang, and J.~Kautz, ``Learning
  superpixels with segmentation-aware affinity loss,'' in \emph{The IEEE
  Conference on Computer Vision and Pattern Recognition (CVPR)}, June 2018.

\bibitem{Yu2017}
F.~Yu, V.~Koltun, and T.~Funkhouser, ``Dilated residual networks,'' in
  \emph{The IEEE Conference on Computer Vision and Pattern Recognition (CVPR)},
  2017.

\bibitem{Bliek2014SolvingMQ}
C.~Bliek, P.~Bonami, and A.~Lodi, ``Solving mixed-integer quadratic programming
  problems with ibm-cplex : a progress report,'' \emph{Proceedings of the RAMP
  Symposium}, 2014.

\bibitem{2018arXiv180202611C}
L.~Chen, Y.~Zhu, G.~Papandreou, F.~Schroff, and H.~Adam, ``Encoder-decoder with
  atrous separable convolution for semantic image segmentation,'' \emph{arXiv
  e-prints}, Feb 2018.

\bibitem{DmitriCVPR19}
D.~Marin, M.~Tang, R.~I.~B. Ayed, and Y.~Boykov, ``Beyond gradient descent for
  regularized segmentation losses,'' in \emph{The IEEE Conference on Computer
  Vision and Pattern Recognition (CVPR)}, 2019.

\end{thebibliography}
